\documentclass[twoside]{article}

\usepackage{hyperref}
\usepackage{amssymb}
\usepackage{thmtools}
\usepackage{amsmath}
\usepackage{amsthm}
\usepackage{amssymb}  
\usepackage{bm}      
\usepackage{booktabs,graphicx}
 \usepackage{multirow}

\usepackage[ruled,vlined,linesnumbered]{algorithm2e}

\newtheorem{proposition}{Proposition}

\newtheorem{assumption}{Assumption}

\usepackage{color, soul}

\usepackage[preprint]{aistats2027}
\usepackage[round]{natbib}

\begin{document}

%
\runningtitle{FedFiOT: Fisher-Weighted Optimal Transport for Model Merging}

%
\runningauthor{Quan Nguyen, Hoang Ngo, Bhakti Chowkwale, Y. Keshtkarjahromi, My T. Thai}
\twocolumn[

\aistatstitle{FedFiOT: Fisher-Weighted Optimal Transport for Model Merging in One-Shot and Few-Round Federated Learning}

\aistatsauthor{ Quan Minh Nguyen$^{1}$, Hoang M. Ngo$^{1}$, Bhakti Chowkwale$^{2}$ \\[2pt]
                \bfseries Yasaman Keshtkarjahromi$^{2}$, My T. Thai$^{1}$ }

\aistatsaddress{ \rule{0pt}{14pt}$^{1}$University of Florida \qquad $^{2}$Seagate Technology }
]

\begin{abstract}
    One-shot and few-round federated learning (FL) reduce communication burden by constructing a global model within one or a few communication rounds, but often achieve lower performance than conventional multi-round FL.  Without frequent synchronization, client models may drift apart, causing neurons that learn similar features to occupy different positions across clients.  At the same time, local data heterogeneity can cause corresponding parameters to differ in their importance to each client’s predictions. Together, these challenges cause direct averaging to combine mismatched neurons and assign equal weight to parameters despite differences in their importance.

    We propose FedFiOT, a Fisher-weighted optimal transport framework for one-shot and few-round FL that leverages Fisher information to quantify parameter importance and guide neuron alignment and model aggregation. FedFiOT jointly optimizes neuron correspondences and global model parameters through alternating minimization of a shared Fisher-weighted objective. Theoretically, we show that FedFiOT’s optimization procedure monotonically decreases the layer-wise merging objective and establish a stationarity bound that quantifies the effects of client heterogeneity and nonuniform Fisher weighting. Experiments across multiple datasets demonstrate consistent improvements over competing baselines in both one-shot and few-round federated learning settings.
\end{abstract}

\section{Introduction}

Federated learning (FL) \cite{mcmahan2017communication} enables collaborative model training without sharing private data, but often requires a large number of communication rounds to achieve good accuracy. This multi-round training paradigm has several drawbacks. First, clients must frequently be available to connect to the server to send and receive model updates. Second, clients must repeatedly expend computational resources to update the global model whenever they participate in training. Third, repeated communication rounds can increase vulnerability to backdoor, model poisoning or inference attacks by allowing adversaries to continuously refine their attacks in response to the evolving global model \cite{gu2022cs,xie2025model,zhu2025fedmia, zhang2023a3fl}. Furthermore, frequent communication rounds can amplify the communication and computational costs of privacy-preserving mechanisms such as secure aggregation \cite{bonawitz2017practical} and homomorphic encryption \cite{zhang2020batchcrypt}, which impose additional overhead on clients in each round.

These limitations have motivated FL approaches that seek to reduce communication frequency. \emph{One-shot federated learning} constructs a global model using only a single round \cite{guha2019one,zhang2022dense,tang2024fusefl}, while \emph{few-round federated learning} allows only a small number of communication rounds \cite{park2021few, jhunjhunwala2024fedfisher}. Few-round FL provides a middle ground between one-shot and conventional multi-round FL, where the goal is to approach the performance of fully iterative FL using as few communication rounds as possible. However, with fewer synchronization opportunities, the quality of each aggregation step becomes more important, especially when clients' data are heterogeneous.

A key challenge in one-shot and few-round FL is \emph{neuron misalignment}. Due to permutation invariance in neural networks, two models can represent the same function while differing only in the ordering of their hidden neurons or channels \cite{ainsworth2022git, entezari2021role}. Consequently, similar features may be represented by different neurons across client models, causing naive parameter averaging to combine mismatched neurons. Neuron-matching methods \cite{liu2022deep,yurochkin2019bayesian, wang2020federated, singh2020model, akash2022wasserstein} address this issue by identifying correspondences between neurons across client models based on the similarity of their weights or learned representations, and then permuting neurons into a common ordering before parameter aggregation.


A complementary problem is determining how parameters from different models should be combined during model fusion. From a Bayesian perspective, model merging can be viewed as finding parameters that maximize the joint likelihood of the models’ posterior distribution. Fisher merging \cite{matena2022merging} adopts this perspective by approximating each model's posterior as a Gaussian whose precision matrix is given by its Fisher information. The merged model is then obtained through a Fisher-weighted average, giving greater influence to parameter values with higher estimated precision. Intuitively, a large Fisher value indicates that changing a parameter can substantially affect the model’s predictions, whereas a small Fisher value suggests that the parameter has relatively little influence on the model’s behavior.

These two challenges are closely coupled. Since Fisher information is a local property of a single parameter value, Fisher-based merging may become less reliable when the models being merged are far apart in parameter space \cite{matena2022merging}. Consequently, Fisher weighting is most effective when corresponding neurons are properly aligned across models. At the same time, neuron alignment should place greater emphasis on correctly matching neurons whose associated parameters have high Fisher information values, since incorrectly matching parameters that strongly influence the model’s predictions can be substantially more harmful than misaligning parameters with low Fisher values. Existing approaches typically consider Fisher-based weighting and neuron alignment as separate problems, potentially limiting their effectiveness in one-shot and few-round federated learning.

To address this gap, we propose FedFiOT, a Fisher-weighted optimal transport framework for one-shot and few-round FL. We summarize our contributions:
\begin{itemize}
    \item We propose FedFiOT and show how neuron alignment and Fisher-weighted aggregation, which prior work treats separately, can be posed as a single optimization over the client-specific neuron correspondences and the global model, using only the local models and Fisher estimates sent by the clients (Sections~\ref{sec:prob_form} and \ref{sec:fisher_weight_agg}).

    \item We formulate neuron alignment as a damped Fisher-weighted optimal transport problem, assigning larger penalties to discrepancies in parameters that are more important to each client’s predictions. We solve the joint alignment and aggregation problem by alternating between computing layer-wise transport plans and updating global model parameters through Fisher-weighted averaging of aligned client parameters (Section~\ref{sec:gen_proc}).

    \item Our theoretical analysis shows that the alternating updates monotonically decrease the layer-wise merging objective, with damping making each aggregation step strongly convex so that the change in the merged weights vanishes at an $O(S^{-1/2})$ rate. We further establish a stationarity bound across communication rounds that quantifies the effects of client heterogeneity, stochastic noise, and nonuniform Fisher weighting (Section~\ref{sec:theory}).

    \item We evaluate FedFiOT on a range of one-shot and few-round FL tasks and show that it gives a consistent improvement in global model accuracy compared to competing baselines involving knowledge distillation, neuron matching or model fusion (Section~\ref{sec:experiments}).
\end{itemize}

\section{Related Works}

\subsection{One-Shot and Few-Round FL}

Existing one-shot federated learning (FL) methods broadly fall into four categories: (i) knowledge distillation, (ii) neuron matching, (iii) model fusion and (iv) ensemble-based methods. Knowledge distillation methods combine client models into an ensemble and transfer its knowledge to a single global model using either server-side auxiliary data \cite{guha2019one, li2020practical, Gong_2021_ICCV} or synthetic data generated by GANs or VAEs \cite{zhang2022dense, heinbaugh2023data}. However, auxiliary data may poorly represent client distributions, while sharing client data or generating synthetic samples can raise privacy concerns, e.g. data reconstruction attack \cite{hitaj2017deep}. Data generation and distillation also increase server computation and require careful hyperparameter tuning, complicating their use in one-shot FL \cite{jhunjhunwala2024fedfisher}. Neuron matching methods \cite{liu2022deep,yurochkin2019bayesian, wang2020federated, singh2020model, akash2022wasserstein} exploit neural networks’ permutation invariance to align corresponding neurons across client models before averaging their weights. Another line of work, known as model fusion, aims to combine the capabilities of multiple existing models into a single model \cite{choshen2022fusing, ilharco2022editing, jin2022dataless, matena2022merging, yadav2023ties}. Although not explicitly developed for one-shot federated learning, some of these methods offer improvement to simple parameter averaging. However, they primarily focus on merging pre-trained models and do not explicitly account for heterogeneity in the data used to train them. FedBEns~\cite{talpini2025fedbens}, an ensemble-based approach, captures multimodal client posteriors through mixtures of Laplace approximations, but requires clients to train and communicate multiple models and retains an ensemble for prediction. Few-round FL relaxes the single-round communication constraint, allowing clients to refine their models over a small number of communication rounds \cite{jhunjhunwala2024fedfisher,park2021few}.

\subsection{Permutation Invariance \& Neuron Alignment in Federated Learning}

Due to permutation invariance, direct parameter averaging can combine mismatched neurons across clients. Neuron alignment methods address this issue by establishing correspondences between neurons before aggregation. PFNM \cite{yurochkin2019bayesian} uses Bayesian nonparametric inference to determine whether a local neuron should match an existing global neuron or introduce a new one, while FedMA \cite{wang2020federated} extends this approach to CNNs and LSTMs by alternating layer-wise matching with local retraining. Other methods formulate neural layers as distributions over neurons and use optimal transport (OT) to establish correspondences before merging \cite{singh2020model} or graph matching that captures dependencies between connected neurons \cite{liu2022deep}. Although these methods mitigate permutation mismatch, alignment alone does not resolve differences in features learned from heterogeneous client data.

\subsection{Model Fusion in Federated Learning}

Most model fusion approaches target pre-trained or fine-tuned models and offer no theoretical guarantees in FL. A closely related work is Fisher merging \cite{matena2022merging}, which approximates each model's posterior as a Gaussian whose precision is given by its diagonal Fisher information and casts merging as maximizing the models' joint posterior likelihood. FedFisher \cite{jhunjhunwala2024fedfisher} extends this idea to one-shot FL and provides a theoretical analysis for two-layer over-parameterized ReLU networks. However, since the Fisher information is a local property of a single parameter value, Fisher-based merging may perform poorly when the models being merged are misaligned and lie far apart in parameter space. 
\section{Proposed Algorithm: FedFiOT}
\label{sec:fedfiot}
In this section, we first introduce the federated model-merging setup and formulate neuron alignment and parameter aggregation as a joint optimization problem. We then present FedFiOT, and describe its application to one-shot and few-round FL.
\subsection{Federated Model-Merging Setup}
\label{sec:prob_form}

We consider an FL system with $K$ clients, where client $k$ holds a private dataset $\mathcal D_k$. All clients share an $L$-layer network architecture. At each communication round, the server broadcasts a global model, which each participating client uses to initialize local training. Client $k$ then returns its updated parameters $\mathbf W_k\in\mathcal W:=\mathbb R^d$ and a diagonal Fisher estimate $\widetilde{\mathbf F}_k\succeq0$ computed from its local data. We describe a single aggregation round and omit the round index for clarity.

The Fisher estimate provides a measure of parameter importance. Specifically, a larger entry indicates greater sensitivity of the client's predictive distribution to changes in the corresponding parameter~\cite{martens2020new,matena2022merging}. Details of the Fisher estimation procedure are provided in Appendix~\ref{appx:fisher_info}. 

Let $m_\ell$ denote the number of neurons in layer $\ell$. For a fully connected layer, let $w_{k,j}^{\ell}\in\mathbb R^{m_{\ell-1}}$ denote the incoming weights of neuron $j$ in client $k$'s model, and let $\widetilde f_{k,j}^{\ell}$ denote the corresponding diagonal Fisher block. We omit biases from the notation. Client $k$'s layer is $\mathbf w_k^\ell$, with neuron weights $\{w_{k,j}^\ell\}_{j=1}^{m_\ell}$. The global parameters at this layer are denoted by $\mathbf w^\ell\in\mathcal W^\ell:=\mathbb R^{m_\ell\times m_{\ell-1}}$, with neuron weights $\{w_i^\ell\}_{i=1}^{m_\ell}$.

Neurons that learn similar features may appear at different positions within the same layer across clients, and their parameters may have different Fisher values. The aggregation problem therefore involves two coupled decisions: (1) identifying neuron correspondences and (2) determining the global parameter values. Our goal is to formulate these decisions jointly while accounting for the parameter importance indicated by the Fisher estimates.

\begin{figure*}[t]
  \centering
  \includegraphics[width=\textwidth]{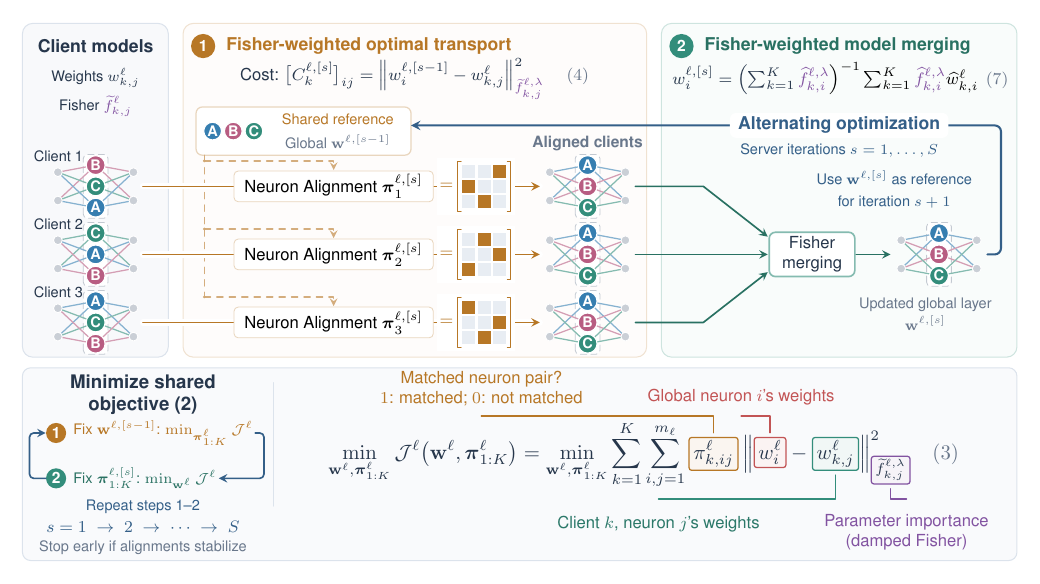}
  \caption{\textbf{Overview of FedFiOT.}
  At each layer, the server minimizes the joint objective $\mathcal J^\ell$ by
  alternating between Fisher-weighted neuron alignment and Fisher-weighted
  model merging. The merged layer serves as the alignment reference for the
  next iteration. This repeats for up to $S$ iterations, or until no client's
  alignment changes.
  Here, $\ell$ indexes the layer and $[s]$ the iteration; $w_i^\ell$ and
  $w_{k,j}^\ell$ denote the weights of global neuron $i$ and of neuron $j$ of client
  $k$. In each permutation matrix, rows are global neurons, columns are client
  neurons, and orange cells mark matched pairs.
  $\widehat w_{k,i}^\ell$ and $\widehat f_{k,i}^{\ell,\lambda}$ denote the weights
and damped Fisher block of the neuron of client~$k$ matched to global neuron~$i$. For illustration, we provide a simple numerical example in
  Appendix~\ref{appx:example}.}
  \label{fig:overview}
\end{figure*}

\subsection{Joint Alignment and Aggregation Objective}
\label{sec:fisher_weight_agg}

We formulate the merge at layer $\ell$ as a joint optimization over neuron correspondences and global parameters. For this formulation, the correspondence between neurons in layer $\ell-1$ is held fixed and determines a common ordering for the incoming coordinates. For the first hidden layer, this ordering is given by the shared input features.

Let $\mathcal P^\ell$ denote the set of $m_\ell\times m_\ell$ permutation matrices. For client $k$, an alignment $\boldsymbol\pi_k^\ell\in\mathcal P^\ell$ specifies the neuron correspondences. Specifically, $\pi_{k,ij}^\ell:=[\boldsymbol\pi_k^\ell]_{ij}=1$ means that client neuron $j$ corresponds to global neuron $i$. We write $\boldsymbol\pi_{1:K}^\ell=(\boldsymbol\pi_1^\ell,\ldots,\boldsymbol\pi_K^\ell)$ for the collection of client alignments. The output-layer alignment is restricted to the identity, since all clients share the same class ordering.

To account for parameter importance, we use the damped Fisher blocks
\begin{equation}
\widetilde f_{k,j}^{\ell,\lambda}
:=
\widetilde f_{k,j}^{\ell}+\lambda I,
\qquad \lambda>0,
\end{equation}
where $\lambda>0$ ensures positive precision even for coordinates with zero estimated Fisher information. Since diagonal Fisher entries often span several orders of magnitude, a few coordinates can dominate a Fisher-weighted average. We therefore compute $\widetilde{\mathbf F}_k$ with each empirical Fisher entry raised to a power $\rho\in(0,1]$ (Appendix~\ref{appx:fisher_info}). This tempering preserves the ordering of parameter importance while compressing its range. To keep the notation light, we absorb $\rho$ into $\widetilde{\mathbf F}_k$ and refer to $\widetilde f_{k,j}^{\ell,\lambda}$ simply as the damped Fisher. Writing $\|x\|_A^2:=x^\top A x$, we define
\begin{equation}
\begin{aligned}
    \mathcal J^\ell
    \bigl(\mathbf w^\ell,\boldsymbol\pi_{1:K}^\ell\bigr) :=
    \sum_{k=1}^{K}\sum_{i,j=1}^{m_\ell}
    \pi_{k,ij}^\ell
    \left\|
        w_i^\ell-w_{k,j}^\ell
    \right\|_{\widetilde f_{k,j}^{\ell,\lambda}}^2.
\end{aligned}
\label{eq:fedfish_objective}
\end{equation}
The resulting layer-merging problem is
\begin{equation}
    \min_{\substack{
        \mathbf w^\ell\in\mathcal W^\ell\\
        \boldsymbol\pi_{1:K}^\ell\in(\mathcal P^\ell)^K
    }}
    \mathcal J^\ell
    \bigl(\mathbf w^\ell,\boldsymbol\pi_{1:K}^\ell\bigr).
\label{eq:fedfiot_layer_objective}
\end{equation}

The objective penalizes discrepancies between matched neurons, with greater weight assigned to coordinates with larger Fisher values. Thus, the same importance estimates guide both the choice of neuron correspondences and the global parameter values.
We propose  FedFiOT, a solution for the layer-merging problem which constructs a merged model
through alternating alignment and aggregation updates.

\subsection{Alternating Alignment and Aggregation}
\label{sec:gen_proc}

We solve~\eqref{eq:fedfiot_layer_objective} by alternating between neuron alignment and parameter aggregation. We index these inner iterations by a bracketed superscript $[s]$. For a fixed layer $\ell$, the global weights $\mathbf w^{\ell,[0]}$ can be initialized from either the broadcast model or a locally trained client model. Each inner iteration then consists of the following two updates.

\paragraph{Fisher-weighted alignment.}
With the global weights fixed at $\mathbf w^{\ell,[s-1]}$, we construct the matching cost
\begin{equation}
    \bigl[C_k^{\ell,[s]}\bigr]_{ij}
    :=
    \left\|
        w_i^{\ell,[s-1]}
        -
        w_{k,j}^{\ell}
    \right\|_{\widetilde f_{k,j}^{\ell,\lambda}}^2,
\label{eq:fisher_ground_cost}
\end{equation}
where $C_k^{\ell,[s]}\in\mathbb R^{m_\ell\times m_\ell}$ collects these costs.
The objective separates across clients, so each alignment is updated independently:
\begin{equation}
    \boldsymbol\pi_k^{\ell,[s]}
    \in
    \operatorname*{arg\,min}_{\boldsymbol\pi\in\mathcal P^\ell}
    \left\langle \boldsymbol\pi,C_k^{\ell,[s]}\right\rangle.
\label{eq:alignment_update}
\end{equation}
This assignment problem is equivalent, up to normalization, to optimal transport between two distributions over the client and global neurons. Specifically, $\boldsymbol\pi_k^{\ell,[s]}/m_\ell$ is an optimal transport plan. We compute the assignment using the Hungarian algorithm~\cite{kuhn1955hungarian}, which takes $\mathcal O(m_\ell^3)$ time. The Fisher-weighted cost makes mismatches in high-importance coordinates more expensive.

\paragraph{Fisher-weighted aggregation.}
\label{sec:fisher_weight_neuron_aggregation}
Once the alignments $\boldsymbol\pi_k^{\ell,[s]}$ are fixed,
we reorder each client's neurons and their Fisher blocks
into the global neuron order. For global neuron $i$, define
\begin{equation}
\begin{aligned}
    \widehat w_{k,i}^{\ell}
    &:=
    \sum_{j=1}^{m_\ell}
    \pi_{k,ij}^{\ell,[s]}w_{k,j}^{\ell},
    \\
    \widehat f_{k,i}^{\ell,\lambda}
    &:=
    \sum_{j=1}^{m_\ell}
    \pi_{k,ij}^{\ell,[s]}
    \widetilde f_{k,j}^{\ell,\lambda}.
\end{aligned}
\label{eq:aligned_neuron_quantities}
\end{equation}
Each row of a permutation matrix contains exactly one
nonzero entry, so these expressions select the client
neuron matched to global neuron $i$ and its corresponding
Fisher block.

With these correspondences fixed, setting the gradient
of~\eqref{eq:fedfish_objective} with respect to
$w_i^\ell$ to zero gives
\begin{equation}
    w_i^{\ell,[s]}
    =
    \left(
        \sum_{k=1}^{K}
        \widehat f_{k,i}^{\ell,\lambda}
    \right)^{-1}
    \sum_{k=1}^{K}
        \widehat f_{k,i}^{\ell,\lambda}
        \widehat w_{k,i}^{\ell}.
\label{eq:closed_form}
\end{equation}
Since the Fisher blocks are diagonal, this update
averages each weight coordinate according to its
corresponding Fisher values. Damping ensures that
the matrix being inverted is positive definite.
We repeat alignment and aggregation until the
alignments stop changing or $S$ iterations are completed.

\paragraph{Merging across layers.}
The server processes layers in order.
Before merging layer $\ell$, it reorders working copies
of the client weights' incoming coordinates and the
corresponding Fisher entries using the final alignment
of layer $\ell-1$. This makes the neuron comparisons
consistent with the preceding layer.
The input alignment is the identity.
At the output layer, the neuron alignment is also fixed to the identity, so only one aggregation update is needed. After all layers are merged, the server compares the full-model merge
objective of the aligned model with that of the identity-aligned merge, and returns the latter
if alignment does not reduce the objective.

\paragraph{One-shot and few-round operation.}
\label{sec:extend_few_round}
The returned model initializes the next communication round. Clients recompute their Fisher estimates after local training, and the server recomputes the alignments during each merge. Setting $T=1$ gives one-shot FL. Larger $T$ repeats the same procedure for few-round FL. For partial participation, all client sums above are restricted to the selected set $\mathcal K^{(t)}$. Algorithm~\ref{alg:fedfiot} summarizes the procedure.

\section{Theoretical Analysis}
\label{sec:theory}

We analyze FedFiOT at two levels. First, we establish stability of the alternating updates within each layer. Second, we bound the average squared gradient of the federated training objective in terms of local training, client heterogeneity, and nonuniform Fisher weighting. Proofs and lemmas are deferred to the appendix.

\subsection{Stability of FedFiOT Aggregation}
\label{sec:fedfiot_stability}

Fix a communication round and a layer $\ell$, with the incoming-coordinate correspondence held fixed. The client weights and Fisher estimates remain unchanged throughout the inner iterations. We write $\mathcal J^{\ell,[s]} :=\mathcal J^\ell(\mathbf w^{\ell,[s]}, \boldsymbol\pi_{1:K}^{\ell,[s]})$ for the objective after iteration $s$ and define the change in the merged layer by
\begin{equation}
    \Delta^{\ell,[s]}
    :=
    \left\|
        \mathbf w^{\ell,[s]}
        -
        \mathbf w^{\ell,[s-1]}
    \right\|_2.
\label{eq:model_change}
\end{equation}
To define the initial objective $\mathcal J^{\ell,[0]}$, we use the broadcast layer $\mathbf w^{\ell,[0]}$ and identity alignments $\boldsymbol\pi_k^{\ell,[0]}=I_{m_\ell}$ for all clients.

For fixed alignments, damping makes the layer objective $2K\lambda$-strongly convex in the global weights. Together with the exact assignment update, this yields the following guarantee.

\begin{restatable}[Stability of the Layer-Wise Merge]
{theorem}{fedfiotstability}
\label{thm:fedfiot_stability}
Suppose $\lambda>0$ and
$\widetilde f_{k,j}^{\ell}\succeq0$
for every client $k$ and neuron $j$.
Let $\mathcal J^{\ell,\star}$ denote the minimum value
of~\eqref{eq:fedfiot_layer_objective}.
Under the exact updates
\eqref{eq:alignment_update} and~\eqref{eq:closed_form},
the objective values satisfy
$\mathcal J^{\ell,[s]}
\le\mathcal J^{\ell,[s-1]}$ for every $s\ge1$.
Moreover, after $S\ge1$ inner updates,
\begin{equation}
    \min_{1\le s\le S}
    \Delta^{\ell,[s]}
    \le
    \sqrt{
        \frac{
            \mathcal J^{\ell,[0]}
            -
            \mathcal J^{\ell,\star}
        }{
            K\lambda S
        }
    }.
\label{eq:fedfiot_stability_bound}
\end{equation}
\end{restatable}

Theorem~\ref{thm:fedfiot_stability} shows that alignment
and aggregation consistently reduce the same layer
objective. The available objective decrease limits
how long the merged weights can continue to change
substantially: among the first $S$ updates, the
smallest consecutive change is bounded at an
$O(S^{-1/2})$ rate. This guarantee concerns the
inner solver for a fixed layer and does not imply
global optimality of the complete network merge.

\subsection{Stationarity of Federated Training}
\label{sec:global_convergence}

We now study FedFiOT across communication rounds and restore the round index, omitted in Section~\ref{sec:fedfiot}, as a superscript $(t)$. The analysis follows the training procedure. Specifically, we first bound the disagreement created by local training, then control its effect on the server's aggregation,
and finally establish a stationarity guarantee.

\paragraph{Local training and optimization assumptions.}
We consider full participation, where all $K$ clients participate in every round. The global objective is $\mathcal L(\mathbf W)
:=K^{-1}\sum_{k=1}^{K}\mathcal L_k(\mathbf W)$.
At round $t$, each client starts from the broadcast model $\mathbf W^{(t)}$ and performs $E$ local SGD steps with learning rate $\eta$. Client $k$ returns
\begin{equation}
    \mathbf W_{k}^{(t)}
    =
    \mathbf W^{(t)}
    -
    \eta\sum_{e=0}^{E-1}\mathbf g_{k,e}^{(t)},
    \label{eq:fl_local_update}
\end{equation}
where $\mathbf g_{k,e}^{(t)}$ is the stochastic gradient evaluated at its $e$-th local iterate.

\begin{assumption}[Federated Optimization]
\label{as:fl_standard}
Each $\mathcal L_k$ is $L_f$-smooth, and the global objective $\mathcal L$ is bounded below by $\mathcal L^\star$.
Stochastic gradients are conditionally unbiased
given the preceding randomness, have conditional
variance at most $\sigma^2$, and have norm at most
$G$ almost surely. Clients use independent fresh mini-batches at each local step. Moreover, for every $\mathbf W$,
\begin{equation}
    \frac{1}{K}\sum_{k=1}^{K}
    \left\|
        \nabla\mathcal L_k(\mathbf W)
        -
        \nabla\mathcal L(\mathbf W)
    \right\|_2^2
    \le \zeta^2.
    \label{eq:fl_heterogeneity}
\end{equation}
\end{assumption}

Here, $\sigma^2$ controls stochastic-gradient noise, while $\zeta^2$ measures disagreement between client gradients at the same model. Smoothness and the gradient bound control the additional drift caused by taking multiple local steps.

\paragraph{Dispersion of the returned client models.}
Before analyzing alignment and aggregation, we
quantify how far the client models separate during
local training. Define their arithmetic mean and
dispersion before alignment as
\begin{equation}
    \begin{aligned}
        \overline{\mathbf W}^{(t+1)}
        &:=
        \frac{1}{K}\sum_{k=1}^{K}
        \mathbf W_{k}^{(t)},
        \\
        V^{(t)}
        &:=
        \frac{1}{K}\sum_{k=1}^{K}
        \left\|
            \mathbf W_{k}^{(t)}
            -
            \overline{\mathbf W}^{(t+1)}
        \right\|_2^2.
    \end{aligned}
    \label{eq:client_dispersion}
\end{equation}
For compactness, let
\begin{equation}
    \mathcal R_{\mathrm{loc}}
    :=
    \zeta^2
    +
    \left(1-\frac{1}{K}\right)\frac{\sigma^2}{E}
    +
    \frac{L_f^2\eta^2(E-1)^2G^2}{4}.
    \label{eq:local_error}
\end{equation}

\begin{restatable}[Dispersion of Local Models]
{lemma}{localdispersion}
\label{lem:local_dispersion}
Under Assumption~\ref{as:fl_standard}, the dispersion
of the returned client models satisfies
\begin{equation}
    \mathbb E V^{(t)}
    \le
    3\eta^2E^2\mathcal R_{\mathrm{loc}}.
    \label{eq:local_dispersion_bound}
\end{equation}
\end{restatable}

The proof decomposes each local update into the gradient at the broadcast model, the drift along the local trajectory, and stochastic-gradient noise. After subtracting the client average, these three components give the three terms in $\mathcal R_{\mathrm{loc}}$. Lemma~\ref{lem:local_dispersion} therefore provides a bound on the disagreement that the server must handle during aggregation.

\paragraph{Effect of FedFiOT aggregation.}
We next compare the model returned by FedFiOT with the arithmetic mean $\overline{\mathbf W}^{(t+1)}$ of the original client uploads. Let $\mathcal P$ denote the full-parameter permutations induced by consistent reordering of hidden neurons, their incoming and outgoing weights, and biases when present, with input and output coordinates fixed. We consider architectures for which these permutations preserve each client loss: $\mathcal L_k(\boldsymbol\Pi\mathbf W)=\mathcal L_k(\mathbf W)$ for every $\boldsymbol\Pi\in\mathcal P$. Proposition~\ref{prop:invariance} in the appendix establishes this property for the fully connected setting.

Let $\widetilde{\mathbf F}_{k}^{(t)}$ denote client $k$'s uploaded diagonal Fisher estimate, and write $\widetilde{\mathbf F}_{k}^{\lambda,(t)} :=\widetilde{\mathbf F}_{k}^{(t)}+\lambda I$ with $\lambda>0$. For full-model alignments $\boldsymbol\Pi_{1:K}=(\boldsymbol\Pi_1,\ldots,\boldsymbol\Pi_K)\in\mathcal P^K$, where $\boldsymbol\Pi_k$ is induced by the layer alignments $\boldsymbol\pi_k^1,\ldots,\boldsymbol\pi_k^{L-1}$, the full-model counterpart of the layer objective in~\eqref{eq:fedfish_objective} is
\begin{equation}
    \mathcal J^{(t)}(\mathbf W,\boldsymbol\Pi_{1:K})
    := 
    \sum_{k=1}^{K}
    \left\|
        \mathbf W-\boldsymbol\Pi_k\mathbf W_{k}^{(t)}
    \right\|_{
        \boldsymbol\Pi_k\widetilde{\mathbf F}_{k}^{\lambda,(t)}
        \boldsymbol\Pi_k^\top
    }^2.
    \label{eq:round_merge_objective}
\end{equation}
This objective sums the layer objectives (i.e., $\mathcal J^{(t)}(\mathbf W,\boldsymbol\Pi_{1:K})
    = \sum_{\ell=1}^{L}
    \mathcal J^\ell
    \bigl(\mathbf w^\ell,\boldsymbol\pi_{1:K}^\ell\bigr)$) with the incoming coordinates and Fisher entries reordered consistently according to the preceding layer's permutation.

Let $\boldsymbol\Pi_{1:K}^{\dagger,(t)}$ denote the permutations associated with the candidate selected by the server, including identity permutations when the identity-aligned candidate is retained. For fixed permutations, the aggregation updates in~\eqref{eq:closed_form} minimize the full-model objective:
\begin{equation}
    \mathbf W^{(t+1)}
    =
    \arg\min_{\mathbf W}
    \mathcal J^{(t)}(\mathbf W,\boldsymbol\Pi_{1:K}^{\dagger,(t)}).
    \label{eq:fedfish_closed_form_round}
\end{equation}
The server's selection rule (Algorithm~\ref{alg:fedfiot}) guarantees
\begin{equation}
    \mathcal J^{(t)}
    \bigl(\mathbf W^{(t+1)},\boldsymbol\Pi_{1:K}^{\dagger,(t)}\bigr)
    \le
    \min_{\mathbf W}
    \mathcal J^{(t)}(\mathbf W,\mathbf I)
    \leq \mathcal J^{(t)}(\overline{\mathbf W}^{(t+1)},\mathbf I)\label{eq:inner_comparison}
\end{equation}
where $\mathbf I$ is identity alignment for all clients. To convert this objective comparison into a bound on model dispersion, we control the range of the Fisher entries.

\begin{assumption}[Fisher Bounds]
\label{as:fisher_bounds}
The undamped Fisher entries satisfy
$0\le[\widetilde{\mathbf F}_{k}^{(t)}]_{cc}\le F_{\max}$ for every client $k$, round $t$, and coordinate $c\in\{1,\ldots,d\}$ of the full parameter vector.
\end{assumption}

After damping, every Fisher entry lies between $\lambda$ and $F_{\max}+\lambda$. Define
\begin{equation}
    \kappa
    :=
    \frac{F_{\max}+\lambda}{\lambda},
    \label{eq:fisher_ratio}
\end{equation}
which controls the conversion from Fisher-weighted cost to Euclidean dispersion.
We also quantify how the aggregation weights differ from uniform averaging. For the selected permutations, let
$$
    \widehat{\mathbf F}_{k}^{\lambda,(t)}
    :=
    \boldsymbol\Pi_{k}^{\dagger,(t)}
    \widetilde{\mathbf F}_{k}^{\lambda,(t)}
    \bigl(\boldsymbol\Pi_{k}^{\dagger,(t)}\bigr)^\top
$$
be client $k$'s damped Fisher in the global coordinate ordering. Define
\begin{equation}
    \begin{aligned}
        \omega_{k,c}^{(t)}
        &:=
        \frac{
            [\widehat{\mathbf F}_{k}^{\lambda,(t)}]_{cc}
        }{
            \sum_{k'=1}^{K}
            [\widehat{\mathbf F}_{k'}^{\lambda,(t)}]_{cc}
        },
        \\
        \beta^{(t)}
        &:=
        \max_{k,c}\left|K\omega_{k,c}^{(t)}-1\right|.
    \end{aligned}
    \label{eq:fisher_heterogeneity}
\end{equation}
Here, $\omega_{k,c}^{(t)}$ is the aggregation weight
assigned to client $k$ at coordinate $c$, and
$\beta^{(t)}=0$ corresponds to uniform averaging of
the aligned models.
Assumption~\ref{as:fisher_bounds} gives
$\beta^{(t)}\le F_{\max}/\lambda$.
Together, permutation invariance, the objective
comparison, and the Fisher bounds yield the
following loss guarantee.

\begin{restatable}[Loss Control after FedFiOT Aggregation]
{lemma}{alignmentloss}
\label{lem:alignment_loss}
Under Assumptions~\ref{as:fl_standard}
and~\ref{as:fisher_bounds}, suppose the selected
permutations preserve every client loss, then we have:
\begin{equation}
    \begin{aligned}
        \mathcal L(\mathbf W^{(t+1)})
        \le{}&
        \mathcal L(\overline{\mathbf W}^{(t+1)})
        +
        \frac{L_f}{2}(1+\kappa)V^{(t)}
        \\
        &+
        \beta^{(t)}\sqrt{\kappa V^{(t)}}
        \left\|
            \nabla\mathcal L(\mathbf W^{(t+1)})
        \right\|_2.
    \end{aligned}
    \label{eq:alignment_loss_bound}
\end{equation}
\end{restatable}

The key to the proof is to combine~\eqref{eq:inner_comparison} with the Fisher bounds to control the aligned models' dispersion around the merge by $\kappa V^{(t)}$. Lemma~\ref{lem:alignment_loss} shows that the loss of the FedFiOT aggregate exceeds that of the plain average $\overline{\mathbf W}^{(t+1)}$ by at most terms controlled by the client dispersion $V^{(t)}$ and the non-uniformity $\beta^{(t)}$ of the aligned Fisher weights.


\paragraph{Stationarity guarantee.}
The arithmetic mean $\overline{\mathbf W}^{(t+1)}$ satisfies a local-SGD descent inequality. Lemma~\ref{lem:alignment_loss} transfers this inequality to the actual FedFiOT output, while Lemma~\ref{lem:local_dispersion} bounds the aggregation error through $\mathcal R_{\mathrm{loc}}$. Together, these results give the following guarantee.

\begin{restatable}[Stationarity of FedFiOT]
{theorem}{fedfiotconvergence}
\label{thm:fedfiot_convergence}
Under Assumptions~\ref{as:fl_standard}
and~\ref{as:fisher_bounds}, if $\beta^{(t)}\le\beta_{\max}$ for every round $t$ and $0<\eta E\le 1/(4L_f)$, then
\begin{equation}
    \begin{aligned}
        &\frac{1}{T+1}\sum_{t=0}^{T}
        \mathbb E
        \left\|
            \nabla\mathcal L(\mathbf W^{(t)})
        \right\|_2^2\le
        \frac{
            4\bigl(
                \mathcal L(\mathbf W^{(0)})
                -
                \mathcal L^\star
            \bigr)
        }{\eta E T}
        \\
        &+
        \frac{8L_f\eta\sigma^2}{K}
        +
        \frac{3}{2}L_f^2\eta^2(E-1)^2G^2
        \\
        &+
        \left[
            6L_f(1+\kappa)\eta E
            +
            12\kappa\beta_{\max}^2
        \right]\mathcal R_{\mathrm{loc}}.
    \end{aligned}
    \label{eq:fedfiot_convergence_bound}
\end{equation}
\end{restatable}

Theorem~\ref{thm:fedfiot_convergence} separates the effects of local optimization and server aggregation. The aggregation contribution depends on the disagreement created by local training and the variation in Fisher weights. Its component proportional to $\eta E$ decreases with the learning rate, whereas persistent nonuniform weighting can leave a residual term.

In particular, for fixed $E$, bounded $\kappa$, and a uniform bound $\beta_{\max}$ independent of $T$, by choosing $\eta=\Theta(T^{-1/2})$, we have: 
\begin{equation}
    \begin{aligned}
        &\frac{1}{T+1}\sum_{t=0}^{T}
        \mathbb E
        \left\|
            \nabla\mathcal L(\mathbf W^{(t)})
        \right\|_2^2
        \\
        &\quad=
        O\!\left(
            T^{-1/2}
            +
            \kappa\beta_{\max}^2
            \left[
                \zeta^2
                +
                \left(1-\frac{1}{K}\right)
                \frac{\sigma^2}{E}
            \right]
        \right).
    \end{aligned}
    \label{eq:fedfiot_rate}
\end{equation}
Uniform Fisher weights in the aligned coordinates
therefore recover a vanishing
$O(T^{-1/2})$ stationarity bound.
With persistent nonuniformity, the guarantee
permits a residual error governed by client
heterogeneity and stochastic noise.
Smaller $\beta_{\max}$ tightens this residual bound,
but a small fixed value does not generally
make it vanish.

\section{Experimental Results}
\label{sec:experiments}


\subsection{Experimental Setup}
\label{sec:experimental_setup}

We evaluate FedFiOT against state-of-the-art (SOTA) neuron matching, knowledge distillation, model fusion, and Bayesian ensemble baselines on CIFAR-10, SVHN \cite{Netzer2011}, and CINIC-10 \cite{darlow2018cinic} for image recognition in a FL setting. Dataset details are provided in Appendix~\ref{appx:more_dataset}.

\paragraph{Baseline Methods}
We compare our method with nine baselines: (i) FedAvg \cite{mcmahan2017communication}, (ii) PFNM \cite{yurochkin2019bayesian}, (iii) OTFusion \cite{singh2020model}, (iv) WB Fusion \cite{akash2022wasserstein}, (v) DENSE \cite{zhang2022dense}, (vi) FisherMerge \cite{matena2022merging}, (vii) RegMean \cite{jin2022dataless}, (viii) FedFisher \cite{jhunjhunwala2024fedfisher}, and (ix) FedBEns \cite{talpini2025fedbens}. FedAvg serves as the standard parameter-averaging baseline. PFNM and OTFusion are neuron-matching methods that align neurons across local models before aggregation. WB Fusion jointly optimizes neuron alignment and aggregation through a Wasserstein barycenter formulation, but does not incorporate Fisher-based parameter importance or provide explicit convergence analysis. DENSE uses data-free knowledge distillation with GAN-generated data at the server. FisherMerge, RegMean, and FedFisher represent model fusion approaches, with FisherMerge and FedFisher using Fisher information to guide aggregation. FedBEns captures multimodal client posteriors using mixtures of Laplace approximations and performs ensemble prediction. We exclude inherently multi-round methods, e.g. FedProx \cite{li2020federated} and SCAFFOLD \cite{karimireddy2020scaffold} since their performance would be similar to FedAvg for one or few round.

\paragraph{Models and Other Details.}
Following prior works \cite{jhunjhunwala2024fedfisher, talpini2025fedbens}, we use the CNN proposed in \cite{wang2020federated}. To simulate data heterogeneity, we partition each dataset among $K$ clients using Dirichlet sampling with parameter $\alpha$, where smaller values indicate greater heterogeneity. We use the same local optimization procedure across all algorithms. Further details, including hyperparameter tuning, are provided in Appendix \ref{appx:Hyperparameter}.

\paragraph{Communication Efficiency}
With a diagonal Fisher approximation, each client uploads $d$ Fisher entries
alongside its $d$ model parameters, doubling the number of scalars it sends
compared to FedAvg. We find that this cost can be reduced to match that of FedAvg using standard quantization techniques without significantly affecting accuracy. \textbf{For all experiments, we ensure that FedFiOT's communication cost matches that of FedAvg}. Further practical considerations are discussed in Appendix~\ref{appx:practical}.

\subsection{One-Shot Performance}
\label{sec:oneshot}
\providecommand{\pmstd}[2]{#1{\scriptsize$\pm$#2}}
\begin{table*}[t]
\centering
\caption{One-shot test accuracy (\%) with the number of clients $K=6$ fixed and the heterogeneity
parameter $\alpha$ varied. Mean $\pm$ std over three seeds. FedFiOT uploads quantized weights and diagonal Fishers.}
\label{tab:oneshot}
\setlength{\tabcolsep}{3.5pt}
\begin{tabular}{lc|ccccccc|c|c}
\toprule
Dataset & $\alpha$ & \texttt{FedAvg} & \texttt{OTFusion}
  & \begin{tabular}[c]{@{}c@{}}\texttt{WB}\\\texttt{Fusion}\end{tabular}
  & \texttt{PFNM} & \texttt{RegMean} & \texttt{DENSE}
  & \begin{tabular}[c]{@{}c@{}}\texttt{Fisher}\\\texttt{Merge}\end{tabular}
  & \begin{tabular}[c]{@{}c@{}}\texttt{FedFisher}\\\texttt{(K-FAC)}\end{tabular}
  & \begin{tabular}[c]{@{}c@{}}\texttt{FedFiOT}\\\texttt{(Ours)}\end{tabular} \\
\midrule
\multirow{2}{*}{SVHN}
  & 0.2 & \pmstd{54.4}{2.8} & \pmstd{54.8}{0.9} & \pmstd{60.8}{1.0} & \pmstd{69.5}{1.3} & \pmstd{61.6}{4.3} & \pmstd{70.6}{0.9} & \pmstd{50.7}{1.3} & \pmstd{72.2}{1.8} &  \pmstd{\textbf{78.9}}{1.9} \\
  & 0.1 & \pmstd{24.3}{2.8} & \pmstd{31.0}{1.5} & \pmstd{34.7}{2.3} & \pmstd{44.1}{2.8} & \pmstd{41.1}{5.0} & \pmstd{48.4}{1.1} & \pmstd{38.4}{3.2} & \pmstd{61.6}{3.6} &  \pmstd{\textbf{66.8}}{0.8} \\
\midrule
\multirow{3}{*}{CIFAR10}
  & 0.2  & \pmstd{34.0}{0.8} & \pmstd{37.0}{1.2} & \pmstd{40.3}{1.2} & \pmstd{46.0}{2.3} & \pmstd{39.2}{1.7} & \pmstd{42.5}{1.7} & \pmstd{31.9}{6.7} & \pmstd{48.7}{1.1} &  \pmstd{\textbf{52.0}}{0.9} \\
  & 0.1  & \pmstd{30.3}{1.0} & \pmstd{32.3}{2.8} & \pmstd{29.0}{1.1} & \pmstd{32.8}{1.3} & \pmstd{30.9}{2.5} & \pmstd{33.3}{1.8} & \pmstd{25.2}{2.8} & \pmstd{38.6}{2.1} &  \pmstd{\textbf{43.3}}{1.8} \\
\midrule
\multirow{2}{*}{CINIC10}
  & 0.2 & \pmstd{31.9}{1.1} & \pmstd{34.4}{1.3} & \pmstd{36.4}{1.1} & \pmstd{39.6}{1.4} & \pmstd{31.6}{1.5} & \pmstd{33.2}{1.2} & \pmstd{26.4}{4.6} & \pmstd{40.5}{2.9} &  \pmstd{\textbf{45.6}}{1.4} \\
  & 0.1 & \pmstd{26.6}{1.2} & \pmstd{29.2}{0.7} & \pmstd{25.4}{0.8} & \pmstd{29.8}{1.2} & \pmstd{25.6}{1.3} & \pmstd{26.0}{0.4} & \pmstd{25.6}{0.1} & \pmstd{32.0}{0.5} &  \pmstd{\textbf{39.6}}{0.3} \\
\bottomrule
\end{tabular}
\end{table*}

\textbf{FedFiOT outperforms baselines:} Table~\ref{tab:oneshot} reports test accuracy after a single communication round. FedFiOT achieves the highest mean accuracy across all six settings, outperforming the strongest baseline, FedFisher K-FAC, by 3.3--7.6 percentage points. Compared with FedAvg, FedFiOT improves accuracy by 13.0--42.5 percentage points without increasing communication costs. FedFiOT also outperforms neuron-matching methods (OTFusion, PFNM, WB Fusion) and Fisher-weighted aggregation methods (Fisher Merge and FedFisher K-FAC). These results support the effectiveness of jointly incorporating neuron alignment and Fisher-weighted aggregation.

We further evaluate FedFiOT under extreme non-i.i.d.\ data distributions (Appendix~\ref{appx:noniid}), where it continues to outperform all baselines.

\textbf{Comparison against ensemble methods.}
Figure~\ref{fig:fedbens} compares FedFiOT with FedBEns, which trains a
mixture of $M = 5$ models on each client, increasing
the computation and communication cost by approximately $5$ times.
Nevertheless, FedFiOT achieves performance comparable to FedBEns, even outperforming it in multiple settings (Appendix \ref{appx:ensemble}).

\textbf{Different client initialization.} We further consider one-shot FL in which clients do not share a common initialization, as in model merging of independently trained networks \citep{singh2020model,ainsworth2022git}. This setting arises when clients hold pre-existing models or the server cannot coordinate the initialization, and it removes the partial alignment that a shared initialization provides. In this setting, the gaps between FedFiOT and the baselines widen considerably (Appendix \ref{appx:diff_init}).

\textbf{Ablation Study.} We ablate the components of FedFiOT, including alignment, the Fisher-weighted matching cost, and Fisher-weighted aggregation and show that removing any of
them reduces accuracy by up to 17--29 points (Appendix~\ref{appx:ablation}).

\subsection{Extension to Few-Round FL}
\label{sec:fewround_extension}
\begin{figure}[t]
\centering
\includegraphics[width=\columnwidth]{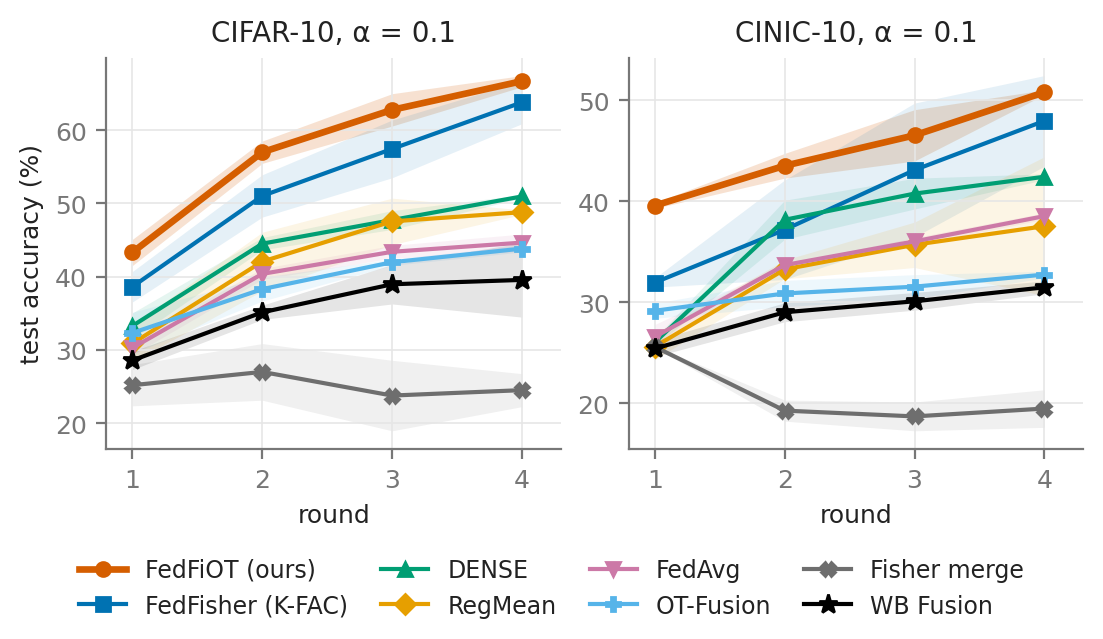}
\caption{Test accuracy (\%) over four rounds of FL with $K=6$ clients and $\alpha=0.1$.}
\label{fig:fewround}
\end{figure}

A natural question is whether FedFiOT provides additional utility in the
few-round setting. Figure~\ref{fig:fewround} reports four rounds with $\alpha=0.1$
and $K=6$. We omit PFNM, which modifies the global architecture and increases client computation and communication in every round, and FedBEns,
which is designed for the one-shot setting. FedFiOT attains the highest accuracy, reaching 66.7\% on CIFAR-10 and 50.8\%  on
CINIC-10 after four rounds. Interestingly, the gap between FedFiOT and FedFisher shrinks over
the rounds. This suggests that after a few rounds of synchronization, clients
start from a shared, well-trained model and their neurons become
increasingly aligned, so the permutation mismatch that FedFiOT corrects
diminishes (Appendix \ref{appx:frac_perm}). FedFiOT's advantage is
therefore most pronounced when communication is limited.

\section{Conclusion}
We propose FedFiOT, a Fisher-weighted optimal transport framework for one-shot
and few-round FL that jointly optimizes neuron alignment and aggregation. Its
alternating updates monotonically decrease the layer-wise merging objective, and we establish a stationarity bound across communication rounds that
quantifies the effects of client heterogeneity, stochastic noise, and
nonuniform Fisher weighting. Empirically, FedFiOT consistently outperforms
one-shot and few-round FL baselines. Future work will extend FedFiOT to richer curvature approximations e.g. K-FAC and make it compatible with secure aggregation.



\bibliography{references}

@inproceedings{mcmahan2017communication,
  title={Communication-efficient learning of deep networks from decentralized data},
  author={McMahan, Brendan and Moore, Eider and Ramage, Daniel and Hampson, Seth and y Arcas, Blaise Aguera},
  booktitle={Artificial intelligence and statistics},
  pages={1273--1282},
  year={2017},
  organization={Pmlr}
}

@article{kuhn1955hungarian,
  title={The Hungarian method for the assignment problem},
  author={Kuhn, Harold W},
  journal={Naval research logistics quarterly},
  volume={2},
  number={1-2},
  pages={83--97},
  year={1955},
  publisher={Wiley Online Library}
}

@article{zhang2023a3fl,
  title={A3fl: Adversarially adaptive backdoor attacks to federated learning},
  author={Zhang, Hangfan and Jia, Jinyuan and Chen, Jinghui and Lin, Lu and Wu, Dinghao},
  journal={Advances in neural information processing systems},
  volume={36},
  pages={61213--61233},
  year={2023}
}

@inproceedings{bonawitz2017practical,
  title={Practical secure aggregation for privacy-preserving machine learning},
  author={Bonawitz, Keith and Ivanov, Vladimir and Kreuter, Ben and Marcedone, Antonio and McMahan, H Brendan and Patel, Sarvar and Ramage, Daniel and Segal, Aaron and Seth, Karn},
  booktitle={proceedings of the 2017 ACM SIGSAC Conference on Computer and Communications Security},
  pages={1175--1191},
  year={2017}
}

@inproceedings{zhang2020batchcrypt,
  title={$\{$BatchCrypt$\}$: Efficient homomorphic encryption for $\{$Cross-Silo$\}$ federated learning},
  author={Zhang, Chengliang and Li, Suyi and Xia, Junzhe and Wang, Wei and Yan, Feng and Liu, Yang},
  booktitle={2020 USENIX annual technical conference (USENIX ATC 20)},
  pages={493--506},
  year={2020}
}

@article{guha2019one,
  title={One-shot federated learning},
  author={Guha, Neel and Talwalkar, Ameet and Smith, Virginia},
  journal={arXiv preprint arXiv:1902.11175},
  year={2019}
}

@article{choshen2022fusing,
  title={Fusing finetuned models for better pretraining},
  author={Choshen, Leshem and Venezian, Elad and Slonim, Noam and Katz, Yoav},
  journal={arXiv preprint arXiv:2204.03044},
  year={2022}
}

@article{ilharco2022editing,
  title={Editing models with task arithmetic},
  author={Ilharco, Gabriel and Ribeiro, Marco Tulio and Wortsman, Mitchell and Gururangan, Suchin and Schmidt, Ludwig and Hajishirzi, Hannaneh and Farhadi, Ali},
  journal={arXiv preprint arXiv:2212.04089},
  year={2022}
}

@article{jin2022dataless,
  title={Dataless knowledge fusion by merging weights of language models},
  author={Jin, Xisen and Ren, Xiang and Preotiuc-Pietro, Daniel and Cheng, Pengxiang},
  journal={arXiv preprint arXiv:2212.09849},
  year={2022}
}

@article{yadav2023ties,
  title={Ties-merging: Resolving interference when merging models},
  author={Yadav, Prateek and Tam, Derek and Choshen, Leshem and Raffel, Colin A and Bansal, Mohit},
  journal={Advances in neural information processing systems},
  volume={36},
  pages={7093--7115},
  year={2023}
}

@article{park2021few,
  title={Few-round learning for federated learning},
  author={Park, Younghyun and Han, Dong-Jun and Kim, Do-Yeon and Seo, Jun and Moon, Jaekyun},
  journal={Advances in Neural Information Processing Systems},
  volume={34},
  pages={28612--28622},
  year={2021}
}

@article{martens2020new,
  title={New insights and perspectives on the natural gradient method},
  author={Martens, James},
  journal={Journal of Machine Learning Research},
  volume={21},
  number={146},
  pages={1--76},
  year={2020}
}

@article{akash2022wasserstein,
  title={Wasserstein barycenter-based model fusion and linear mode connectivity of neural networks},
  author={Akash, Aditya Kumar and Li, Sixu and Trillos, Nicol{\'a}s Garc{\'\i}a},
  journal={arXiv preprint arXiv:2210.06671},
  year={2022}
}

@article{matena2022merging,
  title={Merging models with fisher-weighted averaging},
  author={Matena, Michael S and Raffel, Colin},
  journal={Advances in Neural Information Processing Systems},
  volume={35},
  pages={17703--17716},
  year={2022}
}

@InProceedings{Gong_2021_ICCV,
    author    = {Gong, Xuan and Sharma, Abhishek and Karanam, Srikrishna and Wu, Ziyan and Chen, Terrence and Doermann, David and Innanje, Arun},
    title     = {Ensemble Attention Distillation for Privacy-Preserving Federated Learning},
    booktitle = {Proceedings of the IEEE/CVF International Conference on Computer Vision (ICCV)},
    month     = {October},
    year      = {2021},
    pages     = {15076-15086}
}

@article{zhang2022dense,
  title={Dense: Data-free one-shot federated learning},
  author={Zhang, Jie and Chen, Chen and Li, Bo and Lyu, Lingjuan and Wu, Shuang and Ding, Shouhong and Shen, Chunhua and Wu, Chao},
  journal={Advances in Neural Information Processing Systems},
  volume={35},
  pages={21414--21428},
  year={2022}
}

@inproceedings{hitaj2017deep,
  title={Deep models under the GAN: Information leakage from collaborative deep learning},
  author={Hitaj, Briland and Ateniese, Giuseppe and Perez-Cruz, Fernando},
  booktitle={Proceedings of the 2017 ACM SIGSAC conference on computer and communications security},
  pages={603--618},
  year={2017}
}

@inproceedings{heinbaugh2023data,
  title={Data-free one-shot federated learning under very high statistical heterogeneity},
  author={Heinbaugh, Clare Elizabeth and Luz-Ricca, Emilio and Shao, Huajie},
  booktitle={The Eleventh International Conference on Learning Representations},
  year={2023}
}

@inproceedings{jhunjhunwala2024fedfisher,
  title={Fedfisher: Leveraging fisher information for one-shot federated learning},
  author={Jhunjhunwala, Divyansh and Wang, Shiqiang and Joshi, Gauri},
  booktitle={International Conference on Artificial Intelligence and Statistics},
  pages={1612--1620},
  year={2024},
  organization={PMLR}
}

@article{li2020practical,
  title={Practical one-shot federated learning for cross-silo setting},
  author={Li, Qinbin and He, Bingsheng and Song, Dawn},
  journal={arXiv preprint arXiv:2010.01017},
  year={2020}
}

@article{tang2024fusefl,
  title={Fusefl: One-shot federated learning through the lens of causality with progressive model fusion},
  author={Tang, Zhenheng and Zhang, Yonggang and Dong, Peijie and Cheung, Yiu-ming and Zhou, Amelie C and Han, Bo and Chu, Xiaowen},
  journal={Advances in neural information processing systems},
  volume={37},
  pages={28393--28429},
  year={2024}
}

@article{ainsworth2022git,
  title={Git re-basin: Merging models modulo permutation symmetries},
  author={Ainsworth, Samuel K and Hayase, Jonathan and Srinivasa, Siddhartha},
  journal={arXiv preprint arXiv:2209.04836},
  year={2022}
}

@inproceedings{liu2022deep,
  title={Deep neural network fusion via graph matching with applications to model ensemble and federated learning},
  author={Liu, Chang and Lou, Chenfei and Wang, Runzhong and Xi, Alan Yuhan and Shen, Li and Yan, Junchi},
  booktitle={International Conference on Machine Learning},
  pages={13857--13869},
  year={2022},
  organization={PMLR}
}

@article{wang2020federated,
  title={Federated learning with matched averaging},
  author={Wang, Hongyi and Yurochkin, Mikhail and Sun, Yuekai and Papailiopoulos, Dimitris and Khazaeni, Yasaman},
  journal={arXiv preprint arXiv:2002.06440},
  year={2020}
}

@article{singh2020model,
  title={Model fusion via optimal transport},
  author={Singh, Sidak Pal and Jaggi, Martin},
  journal={Advances in Neural Information Processing Systems},
  volume={33},
  pages={22045--22055},
  year={2020}
}

@inproceedings{yurochkin2019bayesian,
  title={Bayesian nonparametric federated learning of neural networks},
  author={Yurochkin, Mikhail and Agarwal, Mayank and Ghosh, Soumya and Greenewald, Kristjan and Hoang, Nghia and Khazaeni, Yasaman},
  booktitle={International conference on machine learning},
  pages={7252--7261},
  year={2019},
  organization={PMLR}
}

@article{talpini2025fedbens,
  title={FedBEns: One-shot federated learning based on bayesian ensemble},
  author={Talpini, Jacopo and Savi, Marco and Neglia, Giovanni},
  journal={arXiv preprint arXiv:2503.15367},
  year={2025}
}

@article{entezari2021role,
  title={The role of permutation invariance in linear mode connectivity of neural networks},
  author={Entezari, Rahim and Sedghi, Hanie and Saukh, Olga and Neyshabur, Behnam},
  journal={arXiv preprint arXiv:2110.06296},
  year={2021}
}

@inproceedings{zhu2025fedmia,
  title={Fedmia: An effective membership inference attack exploiting" all for one" principle in federated learning},
  author={Zhu, Gongxi and Li, Donghao and Gu, Hanlin and Yao, Yuan and Fan, Lixin and Han, Yuxing},
  booktitle={2025 IEEE/CVF Conference on Computer Vision and Pattern Recognition (CVPR)},
  pages={20643--20653},
  year={2025},
  organization={IEEE}
}

@article{darlow2018cinic,
  title={Cinic-10 is not imagenet or cifar-10},
  author={Darlow, Luke N and Crowley, Elliot J and Antoniou, Antreas and Storkey, Amos J},
  journal={arXiv preprint arXiv:1810.03505},
  year={2018}
}

@inproceedings{Netzer2011,title	= {Reading Digits in Natural Images with Unsupervised Feature Learning},author	= {Yuval Netzer and Tao Wang and Adam Coates and Alessandro Bissacco and Bo Wu and Andrew Y. Ng},year	= {2011},URL	= {http://ufldl.stanford.edu/housenumbers/nips2011_housenumbers.pdf},booktitle	= {NIPS Workshop on Deep Learning and Unsupervised Feature Learning 2011}}

@inproceedings{karimireddy2020scaffold,
  title={Scaffold: Stochastic controlled averaging for federated learning},
  author={Karimireddy, Sai Praneeth and Kale, Satyen and Mohri, Mehryar and Reddi, Sashank and Stich, Sebastian and Suresh, Ananda Theertha},
  booktitle={International conference on machine learning},
  pages={5132--5143},
  year={2020},
  organization={PMLR}
}

@article{li2020federated,
  title={Federated optimization in heterogeneous networks},
  author={Li, Tian and Sahu, Anit Kumar and Zaheer, Manzil and Sanjabi, Maziar and Talwalkar, Ameet and Smith, Virginia},
  journal={Proceedings of Machine learning and systems},
  volume={2},
  pages={429--450},
  year={2020}
}

@inproceedings{xie2025model,
  title={Model poisoning attacks to federated learning via multi-round consistency},
  author={Xie, Yueqi and Fang, Minghong and Gong, Neil Zhenqiang},
  booktitle={2025 IEEE/CVF Conference on Computer Vision and Pattern Recognition (CVPR)},
  pages={15454--15463},
  year={2025},
  organization={IEEE}
}

@article{gu2022cs,
  title={CS-MIA: Membership inference attack based on prediction confidence series in federated learning},
  author={Gu, Yuhao and Bai, Yuebin and Xu, Shubin},
  journal={Journal of Information Security and Applications},
  volume={67},
  pages={103201},
  year={2022},
  publisher={Elsevier}
}
\bibliographystyle{plainnat}

\clearpage
\appendix
\thispagestyle{empty}

\onecolumn
\aistatstitle{Supplementary Materials}
\setcounter{lemma}{0}
\setcounter{theorem}{0}
\section{Proofs}

\fedfiotstability

\begin{proof}
For a fixed merged model $\mathbf w^{\ell,[s-1]}$, the OT alignment step
minimizes the objective over the client permutations. Therefore,
\begin{equation}
    \mathcal J^\ell
    \left(
        \mathbf w^{\ell,[s-1]},
        \boldsymbol\pi_{1:K}^{\ell,[s]}
    \right)
    \le
    \mathcal J^{\ell,[s-1]}.
    \label{eq:alignment_descent}
\end{equation}

Next, fix $\boldsymbol\pi_{1:K}^{\ell,[s]}$.
The merging subproblem is
$2K\lambda$-strongly convex because its Hessian with respect to each $w_i^\ell$ is $2\sum_{k}\widehat f_{k,i}^{\ell,\lambda}\succeq2K\lambda I$, and
$\mathbf w^{\ell,[s]}$ is its minimizer. Hence,
\begin{align}
    &
    \mathcal J^\ell
    \left(
        \mathbf w^{\ell,[s-1]},
        \boldsymbol\pi_{1:K}^{\ell,[s]}
    \right)
    -
    \mathcal J^\ell
    \left(
        \mathbf w^{\ell,[s]},
        \boldsymbol\pi_{1:K}^{\ell,[s]}
    \right)
    \nonumber\\
    &\qquad\ge
    K\lambda
    \left\|
        \mathbf w^{\ell,[s]}
        -
        \mathbf w^{\ell,[s-1]}
    \right\|_2^2.
    \label{eq:merging_descent}
\end{align}

Combining~\eqref{eq:alignment_descent} and \eqref{eq:merging_descent} gives
\begin{equation}
    \mathcal J^{\ell,[s-1]}
    -
    \mathcal J^{\ell,[s]}
    \ge
    K\lambda
    \left(
        \Delta^{\ell,[s]}
    \right)^2.
    \label{eq:iteration_descent}
\end{equation}
Thus, the objective is monotonically non-increasing.

Summing~\eqref{eq:iteration_descent} over $s=1,\ldots,S$ gives
\begin{equation}
    K\lambda
    \sum_{s=1}^{S}
    \left(
        \Delta^{\ell,[s]}
    \right)^2
    \le
    \mathcal J^{\ell,[0]}
    -
    \mathcal J^{\ell,[S]}
    \le
    \mathcal J^{\ell,[0]}
    -
    \mathcal J^{\ell,\star}.
\end{equation}
Since
\begin{equation}
    S
    \min_{1\le s\le S}
    \left(
        \Delta^{\ell,[s]}
    \right)^2
    \le
    \sum_{s=1}^{S}
    \left(
        \Delta^{\ell,[s]}
    \right)^2,
\end{equation}
Eq.~\eqref{eq:fedfiot_stability_bound} follows.
\end{proof}

\localdispersion

\begin{proof}
Let $\mathbf W_{k,e}^{(t)}$ denote the local model of client $k$
immediately before local step $e$. Thus,
\begin{equation}
    \mathbf W_{k,0}^{(t)}
    =
    \mathbf W^{(t)},
    \qquad
    \mathbf W_{k,e+1}^{(t)}
    =
    \mathbf W_{k,e}^{(t)}
    -
    \eta\mathbf g_{k,e}^{(t)},
\end{equation}
so that $\mathbf W_{k}^{(t)}=\mathbf W_{k,E}^{(t)}$.

We decompose the local stochastic gradients into three components. Define
\begin{align}
    \boldsymbol\xi_{k,e}^{(t)}
    &:=
    \mathbf g_{k,e}^{(t)}
    -
    \nabla\mathcal L_k
    \left(
        \mathbf W_{k,e}^{(t)}
    \right),
    \label{eq:local_noise}
    \\
    \mathbf d_{k}^{(t)}
    &:=
    \frac{1}{E}
    \sum_{e=0}^{E-1}
    \left[
        \nabla\mathcal L_k
        \left(
            \mathbf W_{k,e}^{(t)}
        \right)
        -
        \nabla\mathcal L_k(\mathbf W^{(t)})
    \right],
    \label{eq:local_drift}
    \\
    \mathbf z_{k}^{(t)}
    &:=
    \frac{1}{E}
    \sum_{e=0}^{E-1}
    \boldsymbol\xi_{k,e}^{(t)}.
    \label{eq:local_noise_average}
\end{align}
Here, $\mathbf d_{k}^{(t)}$ measures the drift caused by evaluating gradients
along the local trajectory, while $\mathbf z_{k}^{(t)}$ is the average
stochastic-gradient error.

Using the local SGD update, the local change $\Delta_{k}^{(t)}:=\mathbf W_{k}^{(t)}-\mathbf W^{(t)}$ satisfies
\begin{align}
    \Delta_{k}^{(t)}
    &=
    -\eta
    \sum_{e=0}^{E-1}
    \mathbf g_{k,e}^{(t)}
    \nonumber\\
    &=
    -\eta E
    \left[
        \nabla\mathcal L_k(\mathbf W^{(t)})
        +
        \mathbf d_{k}^{(t)}
        +
        \mathbf z_{k}^{(t)}
    \right].
    \label{eq:proof_local_decomposition}
\end{align}

We first control the local drift. Since
$\|\mathbf g_{k,e}^{(t)}\|_2\le G$ almost surely,
\begin{align}
    \left\|
        \mathbf W_{k,e}^{(t)}-\mathbf W^{(t)}
    \right\|_2
    &=
    \left\|
        -\eta
        \sum_{h=0}^{e-1}
        \mathbf g_{k,h}^{(t)}
    \right\|_2
    \nonumber\\
    &\le
    \eta
    \sum_{h=0}^{e-1}
    \left\|
        \mathbf g_{k,h}^{(t)}
    \right\|_2
    \le
    \eta eG.
    \label{eq:local_parameter_drift}
\end{align}
By $L_f$-smoothness of $\mathcal L_k$,
\begin{align}
    \|\mathbf d_{k}^{(t)}\|_2
    &\le
    \frac{1}{E}
    \sum_{e=0}^{E-1}
    \left\|
        \nabla\mathcal L_k
        \left(
            \mathbf W_{k,e}^{(t)}
        \right)
        -
        \nabla\mathcal L_k(\mathbf W^{(t)})
    \right\|_2
    \nonumber\\
    &\le
    \frac{L_f\eta G}{E}
    \sum_{e=0}^{E-1}e
    \nonumber\\
    &=
    \frac{L_f\eta(E-1)G}{2}.
    \label{eq:proof_drift_bound}
\end{align}

We next control the stochastic-gradient error.
Let $\mathcal F^{(t)}$ contain all randomness preceding round $t$, and let
$\mathcal F_{e}^{(t)}$ contain $\mathcal F^{(t)}$ and all randomness preceding local
step $e$. By conditional unbiasedness,
\begin{equation}
    \mathbb E
    \left[
        \boldsymbol\xi_{k,e}^{(t)}
        \mid
        \mathcal F_{e}^{(t)}
    \right]
    =
    \mathbf 0,
\end{equation}
and
\begin{equation}
    \mathbb E
    \left[
        \left\|
            \boldsymbol\xi_{k,e}^{(t)}
        \right\|_2^2
        \mid
        \mathcal F_{e}^{(t)}
    \right]
    \le
    \sigma^2.
\end{equation}

Errors from different local steps have zero cross moments because the earlier
error is measurable with respect to the filtration preceding the later step.
Fresh independent mini-batches across clients also give zero cross moments
between different clients. Therefore,
\begin{equation}
    \mathbb E^{(t)}\mathbf z_{k}^{(t)}
    =
    \mathbf0,
    \qquad
    \mathbb E^{(t)}
    \|\mathbf z_{k}^{(t)}\|_2^2
    \le
    \frac{\sigma^2}{E},
    \label{eq:noise_average_bound}
\end{equation}
where
$\mathbb E^{(t)}[\cdot]
:=
\mathbb E[\cdot\mid\mathcal F^{(t)}]$.

Let
\begin{equation}
    \overline{\mathbf d}^{(t)}
    :=
    \frac1K\sum_{k=1}^{K}\mathbf d_{k}^{(t)},
    \qquad
    \overline{\mathbf z}^{(t)}
    :=
    \frac1K\sum_{k=1}^{K}\mathbf z_{k}^{(t)}.
\end{equation}
Using independence across clients, the cross moments between different clients vanish, so
\begin{equation}
    \mathbb E^{(t)}
    \left\|
        \overline{\mathbf z}^{(t)}
    \right\|_2^2 =\frac{1}{K^2}\sum_{k=1}^{K}\mathbb E^{(t)}\|\mathbf z_{k}^{(t)}\|_2^2
    \le
    \frac{\sigma^2}{KE}.
\end{equation}

The centering identity
\begin{equation}
    \frac1K
    \sum_{k=1}^{K}
    \|\mathbf a_k-\overline{\mathbf a}\|_2^2
    =
    \frac1K
    \sum_{k=1}^{K}
    \|\mathbf a_k\|_2^2
    -
    \|\overline{\mathbf a}\|_2^2
\end{equation}
then gives
\begin{equation}
\begin{aligned}
    \frac1K
    \sum_{k=1}^{K}
    \mathbb E^{(t)}
    \left\|
        \mathbf z_{k}^{(t)}
        -
        \overline{\mathbf z}^{(t)}
    \right\|_2^2
    &\le
    \left(
        1-\frac1K
    \right)
    \frac{\sigma^2}{E}.
\end{aligned}
\label{eq:centered_noise_bound}
\end{equation}
Similarly, by~\eqref{eq:proof_drift_bound},
\begin{align}
    \frac1K
    \sum_{k=1}^{K}
    \left\|
        \mathbf d_{k}^{(t)}
        -
        \overline{\mathbf d}^{(t)}
    \right\|_2^2
    &\le
    \frac1K
    \sum_{k=1}^{K}
    \|\mathbf d_{k}^{(t)}\|_2^2
    \nonumber\\
    &\le
    \frac{
        L_f^2\eta^2(E-1)^2G^2
    }{4}.
    \label{eq:centered_drift_bound}
\end{align}

Averaging~\eqref{eq:proof_local_decomposition} over clients and subtracting
the result from the update of client $k$ gives
\begin{align}
    \mathbf W_{k}^{(t)}
    -
    \overline{\mathbf W}^{(t+1)}
    =
    -\eta E
    \Big[
        &\nabla\mathcal L_k(\mathbf W^{(t)})
        -
        \nabla\mathcal L(\mathbf W^{(t)})
        \nonumber\\
        &+
        \mathbf d_{k}^{(t)}
        -
        \overline{\mathbf d}^{(t)}
        \nonumber\\
        &+
        \mathbf z_{k}^{(t)}
        -
        \overline{\mathbf z}^{(t)}
    \Big].
\end{align}

Applying
\begin{equation}
    \|\mathbf a+\mathbf b+\mathbf c\|_2^2
    \le
    3
    \left(
        \|\mathbf a\|_2^2
        +
        \|\mathbf b\|_2^2
        +
        \|\mathbf c\|_2^2
    \right),
\end{equation}
averaging over clients, and taking conditional expectation gives
\begin{align}
    \mathbb E^{(t)} V^{(t)}
    \le
    3\eta^2E^2
    \Bigg[
        &
        \frac1K
        \sum_{k=1}^{K}
        \left\|
            \nabla\mathcal L_k(\mathbf W^{(t)})
            -
            \nabla\mathcal L(\mathbf W^{(t)})
        \right\|_2^2
        \nonumber\\
        &+
        \frac1K
        \sum_{k=1}^{K}
        \left\|
            \mathbf d_{k}^{(t)}
            -
            \overline{\mathbf d}^{(t)}
        \right\|_2^2
        \nonumber\\
        &+
        \frac1K
        \sum_{k=1}^{K}
        \mathbb E^{(t)}
        \left\|
            \mathbf z_{k}^{(t)}
            -
            \overline{\mathbf z}^{(t)}
        \right\|_2^2
    \Bigg].
\end{align}

Using Assumption~\ref{as:fl_standard},
Eq.~\eqref{eq:centered_noise_bound}, and
Eq.~\eqref{eq:centered_drift_bound},
\begin{equation}
    \mathbb E^{(t)}V^{(t)}
    \le
    3\eta^2E^2
    \left[
        \zeta^2
        +
        \left(
            1-\frac1K
        \right)
        \frac{\sigma^2}{E}
        +
        \frac{
            L_f^2\eta^2(E-1)^2G^2
        }{4}
    \right].
\end{equation}
Taking total expectation proves
Eq.~\eqref{eq:local_dispersion_bound}.
\end{proof}

\begin{proposition}[Permutation invariance]
\label{prop:invariance}
Let $\mathcal P$ denote the set of coordinate permutations of $\mathcal W$
induced by tuples of permutation matrices $(\boldsymbol\pi^1,\dots,\boldsymbol\pi^{L-1})$, $\boldsymbol\pi^\ell\in\mathcal P^\ell$, through
$\mathbf w^{\ell}\mapsto \boldsymbol\pi^\ell\mathbf w^{\ell}(\boldsymbol\pi^{\ell-1})^{\top}$,
$\mathbf b^{\ell}\mapsto \boldsymbol\pi^\ell\mathbf b^{\ell}$, with $\boldsymbol\pi^0=\boldsymbol\pi^L=I$. If every activation commutes with coordinate permutation, then every
$\boldsymbol\Pi\in\mathcal P$ is orthogonal and
$\mathcal{L}_k(\boldsymbol\Pi\mathbf W)=\mathcal{L}_k(\mathbf W)$
for every client $k$ and every $\mathbf W$.
\end{proposition}

\begin{proof}
Each $\boldsymbol\Pi\in\mathcal P$ only reorders the coordinates of $\mathbf W$, so it is a permutation matrix and hence orthogonal. Let $h^0=x$, let $h^\ell=\phi(\mathbf w^\ell h^{\ell-1}+\mathbf b^\ell)$ denote the activations of the original network, and let $h'^\ell$ denote those of the permuted network. If $h'^{\ell-1}=\boldsymbol\pi^{\ell-1}h^{\ell-1}$, then
\begin{equation*}
    h'^\ell
    =\phi\bigl(\boldsymbol\pi^\ell\mathbf w^\ell(\boldsymbol\pi^{\ell-1})^\top\boldsymbol\pi^{\ell-1}h^{\ell-1}+\boldsymbol\pi^\ell\mathbf b^\ell\bigr)
    =\phi\bigl(\boldsymbol\pi^\ell(\mathbf w^\ell h^{\ell-1}+\mathbf b^\ell)\bigr)
    =\boldsymbol\pi^\ell h^\ell,
\end{equation*}
where the last step uses that $\phi$ commutes with coordinate permutations. By induction from $\boldsymbol\pi^0=I$, and since $\boldsymbol\pi^L=I$, both networks produce the same output on every input. Hence $\mathcal L_k(\boldsymbol\Pi\mathbf W)=\mathcal L_k(\mathbf W)$.
\end{proof}

Proposition~\ref{prop:invariance} ensures that FedFiOT can reorder neurons without changing the client objectives. This lets the analysis compare aligned client models with their original counterparts while preserving both
their losses and Euclidean norms.

\alignmentloss

\begin{proof}
Fix a realization of local training, Fisher estimation, and the selected
permutations $\boldsymbol\Pi_{1:K}^{\dagger,(t)}$. For compactness, define
\begin{equation}
    \mathbf X
    :=
    \mathbf W^{(t+1)},
    \qquad
    \mathbf X_k
    :=
    \boldsymbol\Pi_{k}^{\dagger,(t)}
    \mathbf W_{k}^{(t)},
\end{equation}
and let
\begin{equation}
    \overline{\mathbf X}
    :=
    \frac1K
    \sum_{k=1}^{K}
    \mathbf X_k
\end{equation}
be the arithmetic mean of the aligned client models.

We first control how far the aligned client models lie from the FedFiOT
aggregate. By~\eqref{eq:inner_comparison},
\begin{align}
    \mathcal J^{(t)}
    \left(
        \mathbf W^{(t+1)},
        \boldsymbol\Pi_{1:K}^{\dagger,(t)}
    \right)
    &\le
    \min_{\mathbf W}
    \mathcal J^{(t)}(\mathbf W,\mathbf I)
    \nonumber\\
    &\le
    \mathcal J^{(t)}
    \left(
        \overline{\mathbf W}^{(t+1)},
        \mathbf I
    \right).
    \label{eq:objective_comparison_proof}
\end{align}
Since
\begin{equation}
    \widetilde{\mathbf F}_{k}^{\lambda,(t)}
    \preceq
    (F_{\max}+\lambda) I,
\end{equation}
we have
\begin{align}
    \mathcal J^{(t)}
    \left(
        \overline{\mathbf W}^{(t+1)},
        \mathbf I
    \right)
    &\le
    (F_{\max}+\lambda)
    \sum_{k=1}^{K}
    \left\|
        \mathbf W_{k}^{(t)}
        -
        \overline{\mathbf W}^{(t+1)}
    \right\|_2^2
    \nonumber\\
    &=
    K(F_{\max}+\lambda)V^{(t)}.
    \label{eq:identity_objective_bound}
\end{align}

On the other hand, damping and orthogonality of the permutation matrices give
\begin{equation}
    \widehat{\mathbf F}_{k}^{\lambda,(t)}
    \succeq
    \lambda I.
\end{equation}
Therefore, defining the aligned dispersion around the FedFiOT aggregate by
\begin{equation}
    V_{\mathrm{ali}}^{(t)}
    :=
    \frac1K
    \sum_{k=1}^{K}
    \|\mathbf X_k-\mathbf X\|_2^2,
\end{equation}
we obtain
\begin{align}
    V_{\mathrm{ali}}^{(t)}
    &\le
    \frac{
        \mathcal J^{(t)}
        \left(
            \mathbf W^{(t+1)},
            \boldsymbol\Pi_{1:K}^{\dagger,(t)}
        \right)
    }{K\lambda}
    \nonumber\\
    &\le
    \frac{F_{\max}+\lambda}{\lambda}
    V^{(t)}
    =
    \kappa V^{(t)}.
    \label{eq:aligned_spread_bound}
\end{align}

We next control the difference between the Fisher-weighted aggregate
$\mathbf X$ and the arithmetic aligned mean $\overline{\mathbf X}$.
Let
\begin{equation}
    \mathbf A_{k}^{(t)}
    :=
    \operatorname{diag}
    \left(
        \omega_{k,1}^{(t)},
        \ldots,
        \omega_{k,d}^{(t)}
    \right).
\end{equation}
The closed-form FedFiOT update gives
\begin{equation}
    \mathbf X
    =
    \sum_{k=1}^{K}
    \mathbf A_{k}^{(t)}\mathbf X_k,
    \qquad
    \sum_{k=1}^{K}
    \mathbf A_{k}^{(t)}
    =
    I.
    \label{eq:weighted_aligned_mean}
\end{equation}
From the definition of $\beta^{(t)}$,
\begin{equation}
    \left\|
        \mathbf A_{k}^{(t)}
        -
        \frac1K I
    \right\|_{\mathrm{op}}
    \le
    \frac{\beta^{(t)}}{K}.
\end{equation}
Since
\begin{equation}
    \sum_{k=1}^{K}
    \left(
        \mathbf A_{k}^{(t)}
        -
        \frac1K I
    \right)
    =
    \mathbf0,
\end{equation}
we may write
\begin{align}
    \mathbf X-\overline{\mathbf X}
    &=
    \sum_{k=1}^{K}
    \left(
        \mathbf A_{k}^{(t)}
        -
        \frac1K I
    \right)
    \mathbf X_k
    \nonumber\\
    &=
    \sum_{k=1}^{K}
    \left(
        \mathbf A_{k}^{(t)}
        -
        \frac1K I
    \right)
    (\mathbf X_k-\mathbf X).
\end{align}
Hence,
\begin{align}
    \|\mathbf X-\overline{\mathbf X}\|_2
    &\le
    \frac{\beta^{(t)}}{K}
    \sum_{k=1}^{K}
    \|\mathbf X_k-\mathbf X\|_2
    \nonumber\\
    &\le
    \beta^{(t)}
    \left(
        \frac1K
        \sum_{k=1}^{K}
        \|\mathbf X_k-\mathbf X\|_2^2
    \right)^{1/2}
    \nonumber\\
    &=
    \beta^{(t)}
    \sqrt{
        V_{\mathrm{ali}}^{(t)}
    }.
    \label{eq:fisher_reweighting_gap}
\end{align}

We now compare the losses of the aligned and unaligned arithmetic means.
By $L_f$-smoothness,
\begin{align}
    \mathcal L
    \left(
        \mathbf W_{k}^{(t)}
    \right)
    \le{}&
    \mathcal L
    \left(
        \overline{\mathbf W}^{(t+1)}
    \right)
    \nonumber\\
    &+
    \left\langle
        \nabla\mathcal L
        \left(
            \overline{\mathbf W}^{(t+1)}
        \right),
        \mathbf W_{k}^{(t)}
        -
        \overline{\mathbf W}^{(t+1)}
    \right\rangle
    \nonumber\\
    &+
    \frac{L_f}{2}
    \left\|
        \mathbf W_{k}^{(t)}
        -
        \overline{\mathbf W}^{(t+1)}
    \right\|_2^2.
\end{align}
Averaging over clients cancels the linear terms, giving
\begin{equation}
    \frac1K
    \sum_{k=1}^{K}
    \mathcal L
    \left(
        \mathbf W_{k}^{(t)}
    \right)
    \le
    \mathcal L
    \left(
        \overline{\mathbf W}^{(t+1)}
    \right)
    +
    \frac{L_f}{2}V^{(t)}.
    \label{eq:unaligned_mean_loss}
\end{equation}

Similarly, the lower smoothness inequality around
$\overline{\mathbf X}$ gives
\begin{align}
    \mathcal L(\mathbf X_k)
    \ge{}&
    \mathcal L(\overline{\mathbf X})
    +
    \left\langle
        \nabla\mathcal L(\overline{\mathbf X}),
        \mathbf X_k-\overline{\mathbf X}
    \right\rangle
    \nonumber\\
    &-
    \frac{L_f}{2}
    \|\mathbf X_k-\overline{\mathbf X}\|_2^2.
\end{align}
Again averaging cancels the linear terms, and therefore
\begin{equation}
    \mathcal L(\overline{\mathbf X})
    \le
    \frac1K
    \sum_{k=1}^{K}
    \mathcal L(\mathbf X_k)
    +
    \frac{L_f}{2K}
    \sum_{k=1}^{K}
    \|\mathbf X_k-\overline{\mathbf X}\|_2^2.
    \label{eq:aligned_mean_loss}
\end{equation}

By Proposition~\ref{prop:invariance},
\begin{equation}
    \mathcal L(\mathbf X_k)
    =
    \mathcal L
    \left(
        \boldsymbol\Pi_{k}^{\dagger,(t)}
        \mathbf W_{k}^{(t)}
    \right)
    =
    \mathcal L
    \left(
        \mathbf W_{k}^{(t)}
    \right).
\end{equation}
Combining this with
Eqs.~\eqref{eq:unaligned_mean_loss} and
\eqref{eq:aligned_mean_loss} gives
\begin{align}
    \mathcal L(\overline{\mathbf X})
    \le{}&
    \mathcal L
    \left(
        \overline{\mathbf W}^{(t+1)}
    \right)
    +
    \frac{L_f}{2}V^{(t)}
    \nonumber\\
    &+
    \frac{L_f}{2K}
    \sum_{k=1}^{K}
    \|\mathbf X_k-\overline{\mathbf X}\|_2^2.
    \label{eq:comparison_of_means}
\end{align}

Finally, apply the lower smoothness inequality at $\mathbf X$ with
$\overline{\mathbf X}$ and rearrange:
\begin{align}
    \mathcal L(\mathbf X)
    \le{}&
    \mathcal L(\overline{\mathbf X})
    +
    \left\langle
        \nabla\mathcal L(\mathbf X),
        \mathbf X-\overline{\mathbf X}
    \right\rangle
    +
    \frac{L_f}{2}
    \|\mathbf X-\overline{\mathbf X}\|_2^2.
\end{align}
The variance decomposition
\begin{equation}
    V_{\mathrm{ali}}^{(t)}
    =
    \frac1K
    \sum_{k=1}^{K}
    \|\mathbf X_k-\overline{\mathbf X}\|_2^2
    +
    \|\mathbf X-\overline{\mathbf X}\|_2^2
\end{equation}
therefore yields
\begin{align}
    \mathcal L(\mathbf X)
    \le{}&
    \mathcal L
    \left(
        \overline{\mathbf W}^{(t+1)}
    \right)
    +
    \frac{L_f}{2}
    \left(
        V^{(t)}+V_{\mathrm{ali}}^{(t)}
    \right)
    \nonumber\\
    &+
    \left\langle
        \nabla\mathcal L(\mathbf X),
        \mathbf X-\overline{\mathbf X}
    \right\rangle.
\end{align}
Using Cauchy--Schwarz and
Eq.~\eqref{eq:fisher_reweighting_gap},
\begin{align}
    \mathcal L(\mathbf X)
    \le{}&
    \mathcal L
    \left(
        \overline{\mathbf W}^{(t+1)}
    \right)
    +
    \frac{L_f}{2}
    \left(
        V^{(t)}+V_{\mathrm{ali}}^{(t)}
    \right)
    \nonumber\\
    &+
    \beta^{(t)}
    \sqrt{
        V_{\mathrm{ali}}^{(t)}
    }
    \|\nabla\mathcal L(\mathbf X)\|_2.
\end{align}
Finally,
$V_{\mathrm{ali}}^{(t)}\le\kappa V^{(t)}$
from~\eqref{eq:aligned_spread_bound} and
$\mathbf X=\mathbf W^{(t+1)}$, so
\begin{align}
    \mathcal L(\mathbf W^{(t+1)})
    \le{}&
    \mathcal L(\overline{\mathbf W}^{(t+1)})
    +
    \frac{L_f}{2}(1+\kappa)V^{(t)}
    \nonumber\\
    &+
    \beta^{(t)}
    \sqrt{\kappa V^{(t)}}
    \left\|
        \nabla\mathcal L(\mathbf W^{(t+1)})
    \right\|_2.
\end{align}
This proves Eq.~\eqref{eq:alignment_loss_bound}.
\end{proof}

\fedfiotconvergence

\begin{proof}
For compactness, define
\begin{equation}
    \gamma:=\eta E,
    \qquad
    D:=
    \frac{L_f\eta(E-1)G}{2}.
    \label{eq:gamma_D}
\end{equation}
Let $\mathcal F^{(t)}$ contain all randomness preceding communication round $t$,
and write
\begin{equation}
    \mathbb E^{(t)}[\cdot]
    :=
    \mathbb E[\cdot\mid\mathcal F^{(t)}].
\end{equation}

We first establish descent for the arithmetic average of the locally trained
models. Averaging the decomposition
in~\eqref{eq:proof_local_decomposition} over clients gives
\begin{equation}
    \overline{\mathbf W}^{(t+1)}
    =
    \mathbf W^{(t)}
    -
    \gamma
    \left[
        \nabla\mathcal L(\mathbf W^{(t)})
        +
        \overline{\mathbf d}^{(t)}
        +
        \overline{\mathbf z}^{(t)}
    \right].
    \label{eq:mean_update}
\end{equation}
From the proof of Lemma~\ref{lem:local_dispersion},
\begin{equation}
    \|\overline{\mathbf d}^{(t)}\|_2
    \le
    D,
    \qquad
    \mathbb E^{(t)}\overline{\mathbf z}^{(t)}
    =
    \mathbf0,
    \qquad
    \mathbb E^{(t)}
    \|\overline{\mathbf z}^{(t)}\|_2^2
    \le
    \frac{\sigma^2}{KE}.
    \label{eq:mean_error_bounds}
\end{equation}

By $L_f$-smoothness of $\mathcal L$,
\begin{align}
    \mathcal L
    \left(
        \overline{\mathbf W}^{(t+1)}
    \right)
    \le{}&
    \mathcal L(\mathbf W^{(t)})
    -
    \gamma
    \|\nabla\mathcal L(\mathbf W^{(t)})\|_2^2
    \nonumber\\
    &-
    \gamma
    \left\langle
        \nabla\mathcal L(\mathbf W^{(t)}),
        \overline{\mathbf d}^{(t)}
        +
        \overline{\mathbf z}^{(t)}
    \right\rangle
    \nonumber\\
    &+
    \frac{L_f\gamma^2}{2}
    \left\|
        \nabla\mathcal L(\mathbf W^{(t)})
        +
        \overline{\mathbf d}^{(t)}
        +
        \overline{\mathbf z}^{(t)}
    \right\|_2^2.
    \label{eq:mean_smoothness}
\end{align}

Since $\nabla\mathcal L(\mathbf W^{(t)})$ is
$\mathcal F^{(t)}$-measurable and
$\mathbb E^{(t)}\overline{\mathbf z}^{(t)}=\mathbf0$,
\begin{equation}
    \mathbb E^{(t)}
    \left\langle
        \nabla\mathcal L(\mathbf W^{(t)}),
        \overline{\mathbf z}^{(t)}
    \right\rangle
    =
    0.
\end{equation}
For the local-drift term, Young's inequality gives
\begin{align}
    -
    \left\langle
        \nabla\mathcal L(\mathbf W^{(t)}),
        \overline{\mathbf d}^{(t)}
    \right\rangle
    &\le
    D
    \|\nabla\mathcal L(\mathbf W^{(t)})\|_2
    \nonumber\\
    &\le
    \frac14
    \|\nabla\mathcal L(\mathbf W^{(t)})\|_2^2
    +
    D^2.
    \label{eq:drift_young}
\end{align}
Moreover,
\begin{align}
    &
    \left\|
        \nabla\mathcal L(\mathbf W^{(t)})
        +
        \overline{\mathbf d}^{(t)}
        +
        \overline{\mathbf z}^{(t)}
    \right\|_2^2
    \nonumber\\
    &\qquad\le
    2
    \|\nabla\mathcal L(\mathbf W^{(t)})\|_2^2
    +
    2
    \|\overline{\mathbf d}^{(t)}+\overline{\mathbf z}^{(t)}\|_2^2
    \nonumber\\
    &\qquad\le
    2
    \|\nabla\mathcal L(\mathbf W^{(t)})\|_2^2
    +
    4D^2
    +
    4
    \|\overline{\mathbf z}^{(t)}\|_2^2.
    \label{eq:mean_quadratic_bound}
\end{align}

Substituting
Eqs.~\eqref{eq:drift_young}--\eqref{eq:mean_quadratic_bound}
into~\eqref{eq:mean_smoothness} and taking conditional expectation yields
\begin{align}
    \mathbb E^{(t)}
    \mathcal L
    \left(
        \overline{\mathbf W}^{(t+1)}
    \right)
    \le{}&
    \mathcal L(\mathbf W^{(t)})
    -
    \gamma
    \left(
        \frac34-L_f\gamma
    \right)
    \|\nabla\mathcal L(\mathbf W^{(t)})\|_2^2
    \nonumber\\
    &+
    \gamma
    (1+2L_f\gamma)D^2
    +
    \frac{
        2L_f\gamma^2\sigma^2
    }{KE}.
    \label{eq:mean_descent_pre}
\end{align}

Since
\begin{equation}
    L_f\gamma
    =
    L_f\eta E
    \le
    \frac14,
\end{equation}
we have
\begin{equation}
    \frac34-L_f\gamma
    \ge
    \frac12,
    \qquad
    1+2L_f\gamma
    \le
    \frac32.
\end{equation}
Thus,
\begin{equation}
\begin{aligned}
    \mathbb E^{(t)}
    \mathcal L
    \left(
        \overline{\mathbf W}^{(t+1)}
    \right)
    \le{}&
    \mathcal L(\mathbf W^{(t)})
    -
    \frac{\gamma}{2}
    \|\nabla\mathcal L(\mathbf W^{(t)})\|_2^2
    \\
    &+
    \frac{3\gamma}{2}D^2
    +
    \frac{
        2L_f\gamma^2\sigma^2
    }{KE}.
\end{aligned}
\label{eq:mean_descent}
\end{equation}

We next transfer this descent inequality to the actual FedFiOT model.
By Lemma~\ref{lem:alignment_loss},
\begin{align}
    \mathcal L(\mathbf W^{(t+1)})
    \le{}&
    \mathcal L
    \left(
        \overline{\mathbf W}^{(t+1)}
    \right)
    +
    \frac{L_f}{2}(1+\kappa)V^{(t)}
    \nonumber\\
    &+
    \beta^{(t)}
    \sqrt{\kappa V^{(t)}}
    \|\nabla\mathcal L(\mathbf W^{(t+1)})\|_2.
\end{align}
Since $\beta^{(t)}\le\beta_{\max}$, Young's inequality gives
\begin{align}
    &
    \beta^{(t)}
    \sqrt{\kappa V^{(t)}}
    \|\nabla\mathcal L(\mathbf W^{(t+1)})\|_2
    \nonumber\\
    &\qquad\le
    \frac{\gamma}{4}
    \|\nabla\mathcal L(\mathbf W^{(t+1)})\|_2^2
    +
    \frac{
        \kappa\beta_{\max}^2
    }{\gamma}
    V^{(t)}.
    \label{eq:aggregation_young}
\end{align}
Therefore,
\begin{equation}
\begin{aligned}
    &
    \mathcal L(\mathbf W^{(t+1)})
    -
    \frac{\gamma}{4}
    \|\nabla\mathcal L(\mathbf W^{(t+1)})\|_2^2
    \\
    &\qquad\le
    \mathcal L
    \left(
        \overline{\mathbf W}^{(t+1)}
    \right)
    +
    \left[
        \frac{L_f}{2}(1+\kappa)
        +
        \frac{\kappa\beta_{\max}^2}{\gamma}
    \right]V^{(t)}.
\end{aligned}
\label{eq:aggregation_descent}
\end{equation}

To telescope this inequality, define the potential
\begin{equation}
    \Phi^{(t)}
    :=
    \mathcal L(\mathbf W^{(t)})
    -
    \frac{\gamma}{4}
    \|\nabla\mathcal L(\mathbf W^{(t)})\|_2^2.
    \label{eq:potential}
\end{equation}
Combining
Eqs.~\eqref{eq:mean_descent} and~\eqref{eq:aggregation_descent} gives
\begin{equation}
\begin{aligned}
    \mathbb E^{(t)}\Phi^{(t+1)}
    \le{}&
    \Phi^{(t)}
    -
    \frac{\gamma}{4}
    \|\nabla\mathcal L(\mathbf W^{(t)})\|_2^2
    \\
    &+
    \frac{3\gamma}{2}D^2
    +
    \frac{
        2L_f\gamma^2\sigma^2
    }{KE}
    \\
    &+
    \left[
        \frac{L_f}{2}(1+\kappa)
        +
        \frac{\kappa\beta_{\max}^2}{\gamma}
    \right]
    \mathbb E^{(t)}V^{(t)}.
\end{aligned}
\label{eq:potential_descent}
\end{equation}

It remains to verify that $\Phi^{(t)}$ is bounded below. By smoothness,
\begin{align}
    \mathcal L^\star
    &\le
    \mathcal L
    \left(
        \mathbf W
        -
        \frac1{L_f}
        \nabla\mathcal L(\mathbf W)
    \right)
    \nonumber\\
    &\le
    \mathcal L(\mathbf W)
    -
    \frac1{2L_f}
    \|\nabla\mathcal L(\mathbf W)\|_2^2.
\end{align}
Hence,
\begin{equation}
    \|\nabla\mathcal L(\mathbf W)\|_2^2
    \le
    2L_f
    \left(
        \mathcal L(\mathbf W)-\mathcal L^\star
    \right).
    \label{eq:gradient_function_gap}
\end{equation}
Consequently,
\begin{align}
    \Phi^{(t)}-\mathcal L^\star
    &=
    \mathcal L(\mathbf W^{(t)})-\mathcal L^\star
    -
    \frac{\gamma}{4}
    \|\nabla\mathcal L(\mathbf W^{(t)})\|_2^2
    \nonumber\\
    &\ge
    \left(
        1-\frac{L_f\gamma}{2}
    \right)
    \left(
        \mathcal L(\mathbf W^{(t)})-\mathcal L^\star
    \right)
    \ge
    0.
    \label{eq:potential_lower_bound}
\end{align}
Thus $\Phi^{(t)}\ge\mathcal L^\star$, while
$\Phi^{(0)}\le\mathcal L(\mathbf W^{(0)})$.

Moreover, by \eqref{eq:gradient_function_gap} and $L_f\gamma\le1/4$,
\begin{equation}
    \Phi^{(T)}-\mathcal L^\star
    \ge\Bigl(\frac{1}{2L_f}-\frac{\gamma}{4}\Bigr)\|\nabla\mathcal L(\mathbf W^{(T)})\|_2^2
    \ge\frac{\gamma}{4}\|\nabla\mathcal L(\mathbf W^{(T)})\|_2^2.
    \label{eq:potential_final_gradient}
\end{equation}

Taking full expectation in~\eqref{eq:potential_descent}, summing over
$t=0,\ldots,T-1$, telescoping, and lower-bounding $\mathbb E\Phi^{(T)}$ by
\eqref{eq:potential_final_gradient}, we have:
\begin{align}
    \frac{\gamma}{4}
    \sum_{t=0}^{T}
    \mathbb E
    \|\nabla\mathcal L(\mathbf W^{(t)})\|_2^2
    \le{}&
    \mathcal L(\mathbf W^{(0)})
    -
    \mathcal L^\star
    \nonumber\\
    &+
    T
    \left[
        \frac{3\gamma}{2}D^2
        +
        \frac{
            2L_f\gamma^2\sigma^2
        }{KE}
    \right]
    \nonumber\\
    &+
    \left[
        \frac{L_f}{2}(1+\kappa)
        +
        \frac{\kappa\beta_{\max}^2}{\gamma}
    \right]
    \sum_{t=0}^{T-1}
    \mathbb EV^{(t)}.
    \label{eq:telescope}
\end{align}

Dividing by $\gamma(T+1)/4$ and using $T/(T+1)\le1$ yields
\begin{align}
    \frac1{T+1}
    \sum_{t=0}^{T}
    \mathbb E
    \|\nabla\mathcal L(\mathbf W^{(t)})\|_2^2
    \le{}&
    \frac{
        4\bigl(
            \mathcal L(\mathbf W^{(0)})-\mathcal L^\star
        \bigr)
    }{\gamma T}
    +
    6D^2
    +
    \frac{
        8L_f\gamma\sigma^2
    }{KE}
    \nonumber\\
    &+
    \left[
        \frac{
            2L_f(1+\kappa)
        }{\gamma}
        +
        \frac{
            4\kappa\beta_{\max}^2
        }{\gamma^2}
    \right]
    \frac1T
    \sum_{t=0}^{T-1}
    \mathbb EV^{(t)}.
    \label{eq:before_dispersion}
\end{align}

By Lemma~\ref{lem:local_dispersion},
\begin{equation}
    \frac1T
    \sum_{t=0}^{T-1}
    \mathbb EV^{(t)}
    \le
    3\gamma^2
    \mathcal R_{\mathrm{loc}}.
    \label{eq:average_dispersion}
\end{equation}
Substituting this into~\eqref{eq:before_dispersion} gives
\begin{align}
    \frac1{T+1}
    \sum_{t=0}^{T}
    \mathbb E
    \|\nabla\mathcal L(\mathbf W^{(t)})\|_2^2
    \le{}&
    \frac{
        4\bigl(
            \mathcal L(\mathbf W^{(0)})-\mathcal L^\star
        \bigr)
    }{\gamma T}
    +
    6D^2
    +
    \frac{
        8L_f\gamma\sigma^2
    }{KE}
    \nonumber\\
    &+
    \left[
        6L_f(1+\kappa)\gamma
        +
        12\kappa\beta_{\max}^2
    \right]
    \mathcal R_{\mathrm{loc}}.
    \label{eq:before_substitution}
\end{align}

Finally,
\begin{equation}
    \gamma=\eta E,
    \qquad
    D^2
    =
    \frac{
        L_f^2\eta^2(E-1)^2G^2
    }{4},
\end{equation}
so
\begin{equation}
    6D^2
    =
    \frac32
    L_f^2\eta^2(E-1)^2G^2
\end{equation}
and
\begin{equation}
    \frac{
        8L_f\gamma\sigma^2
    }{KE}
    =
    \frac{
        8L_f\eta\sigma^2
    }{K}.
\end{equation}
Substituting these expressions into
Eq.~\eqref{eq:before_substitution} gives exactly
Eq.~\eqref{eq:fedfiot_convergence_bound}.
\end{proof}
\section{Fisher Information Matrix and Fisher Merging}
\label{appx:fisher_info}

The Fisher information matrix measures the sensitivity of a
model's predictive distribution to changes in its parameters \cite{jhunjhunwala2024fedfisher, matena2022merging}.
For client $k$, let $p_{\mathbf W_k}(y\mid x)$ denote the
predictive distribution of its locally trained model. Its
Fisher information matrix is
\begin{equation}
\mathbf F_k
=
\frac{1}{N_k}\sum_{n=1}^{N_k}
\mathbb E_{y\sim p_{\mathbf W_k}(\cdot\mid x_{k,n})}
\left[
\mathbf s_{k,n}(y)\mathbf s_{k,n}(y)^\top
\right],
\end{equation}
where $N_k$ is the number of local examples used for estimation,
$x_{k,n}$ is the $n$-th input, and
\begin{equation}
\mathbf s_{k,n}(y)
=
\left.
\nabla_{\mathbf W}\log p_{\mathbf W}(y\mid x_{k,n})
\right|_{\mathbf W=\mathbf W_k}
\end{equation}
is the gradient of the log predictive probability evaluated at
$\mathbf W_k$. Larger diagonal Fisher entries indicate greater
sensitivity to changes in the corresponding parameters.

To avoid storing the full $d\times d$ matrix, we use a diagonal
empirical Fisher estimate computed from observed labels and
tempered by an exponent $\rho\in(0,1]$:
\begin{equation}
[\widetilde{\mathbf F}_k]_{cc}
=
\left(
\frac{1}{N_k}\sum_{n=1}^{N_k}
[\mathbf s_{k,n}(y_{k,n})]_c^2
\right)^{\rho},
\qquad c=1,\ldots,d,
\end{equation}
where $y_{k,n}$ is the observed label and $c$ indexes a parameter.
This estimate uses squared per-example gradients in place of
the expectation over model-generated labels and requires only
$O(d)$ storage. Setting $\rho=1$ recovers the standard diagonal
empirical Fisher. Smaller $\rho$ compresses the range of the entries
(Section~\ref{sec:fisher_weight_agg}).

\section{Additional Experimental Details}
\label{appx:experiment_detail_additional}
\subsection{Datasets}
\label{appx:more_dataset}

We evaluate on three 10-class image-classification benchmarks. CIFAR-10 contains
natural images of objects and animals. SVHN~\cite{Netzer2011} contains
digits cropped from street-view house numbers. CINIC-10~\cite{darlow2018cinic}
extends CIFAR-10 with downsampled ImageNet images of the same ten classes,
which makes it more diverse and harder. We use the standard training and test
splits. All images are $32\times32$ RGB and are normalized per channel using standard
dataset-specific mean and standard deviation values.

\paragraph{Client partitioning.}
We distribute each training set among $K$ clients using a Dirichlet label
split. Each client $k$ draws a class-preference vector
$q_k\sim\operatorname{Dirichlet}(\alpha\mathbf 1_{10})$, and the examples of
each class $c$ are divided among clients in proportion to
$q_{k,c}/\sum_{k'}q_{k',c}$. This produces heterogeneous label distributions
and unequal client dataset sizes, where smaller $\alpha$ concentrates each client on
fewer classes. The partition is generated with a fixed seed and shared by all
methods, so every method sees identical client data. Test accuracy is reported
on the full, class-balanced test set.

\paragraph{Server validation set.}
We reserve 500 examples as a server-side validation set. These examples are
removed from the clients' data and are therefore excluded from both local
training and Fisher estimation. The same validation set is used for
hyperparameter tuning for all methods
(Appendix~\ref{appx:Hyperparameter}).

\subsection{Training Details and Hyperparameter Tuning}
\label{appx:Hyperparameter}

\paragraph{Local training.}
All clients start from a shared Xavier-uniform initialization and train with
SGD (momentum $0.9$, weight decay $2\times10^{-3}$, batch size $64$) and an
initial learning rate of $0.02$ with cosine annealing over the local epochs.
We apply a random $32\times32$ crop with 4-pixel padding and a random horizontal
flip to each minibatch. FedBEns trains its own mixture components
as described below.

\paragraph{Model selection.}
For every method, setting, and seed, we sweep the number of local epochs over
$\{25,50,\ldots,200\}$ and select the value with the highest accuracy on the
500-example server validation set (Appendix~\ref{appx:more_dataset}). This is to ensure that every method is compared at its best local training
length, since longer local training improves each client model but also
increases client drift, and methods differ in how well they tolerate this
drift. Method-specific hyperparameters are selected on the same validation set. We report the mean and sample standard deviation over three training seeds on a fixed data partition. The seeds are shared by all methods within a setting.

\paragraph{FedFiOT.}
Each client computes the diagonal empirical Fisher by averaging squared
per-example gradients of the observed-label log-likelihood over its entire
local dataset (Appendix~\ref{appx:fisher_info}). We sweep the tempering exponent
over $\rho\in\{0.2,0.3,0.4,0.5,0.6,0.7\}$ and the damping over
$\lambda\in\{10^{-3},10^{-4},10^{-5}\}$, and select the pair with the highest
server validation accuracy. All candidates are computed from the same
uploads, so the sweep requires no additional communication. In the one-shot setting, the alignment reference $\mathbf w^{\ell,[0]}$ is the trained model of the client with the most data, since the broadcast model is the random initialization and its neurons encode no learned features to match
against. In later rounds, the sweep additionally selects
the reference from this model and the broadcast model. Each hidden-layer
assignment is solved exactly with the Hungarian algorithm, with at most $S=8$
alternating updates per layer and early stopping once all assignments are
unchanged. Clients upload weights and Fisher entries quantized layer-wise to
16 bits.

\paragraph{Baselines.}
\begin{itemize}

    \item \textbf{FedFisher (K-FAC)} has each client upload a K-FAC
    approximation of its Fisher, compressed by low-rank SVD with 8-bit factors,
    together with its weights quantized to 16 bits. Starting from the
    size-weighted average of the client models, the server minimizes the
    Fisher-weighted distance to the client models using 2{,}000 steps of an
    adaptive momentum optimizer with learning rate $0.01$. Every 100 steps, it
    evaluates the current model on the validation set and keeps the checkpoint
    with the highest validation accuracy.
    \item \textbf{RegMean} has each client upload the Gram matrix of the inputs
    to each fully connected layer, compressed in the same way as the FedFisher
    factors. The server merges each fully connected layer in closed form by
    solving the least-squares problem that best reproduces all clients' layer
    outputs, and averages the convolutional layers by client data size. The
    off-diagonal entries of the Gram matrices are scaled by
    $\alpha\in\{0.1,0.9\}$, selected on validation data.
    \item \textbf{DENSE} distills the ensemble of client models into a single
    model, starting from the FedAvg model. Over 20 stages, a generator is
    trained for 30 Adam steps to synthesize a batch of 256 images on which the
    ensemble is confident but the global model disagrees with it. These images
    are added to a replay pool, and the global model is trained to match the
    ensemble's predictions on the pool (KL divergence, temperature 20, SGD with
    learning rate $0.01$). After each stage, the model is evaluated on the
    validation set, and the stage with the highest validation accuracy is kept.

    \item \textbf{FedBEns} trains five models per client, each from one of five
    random initializations shared across clients, and fits a Kronecker-factored
    Laplace approximation (temperature $0.1$) around each of them. Each client's
    posterior is therefore a mixture of five Gaussians. The server multiplies the
    client posteriors and searches for modes of the product: for each of the five
    initializations, it starts from the coordinate-wise median of the client models
    trained from that initialization and maximizes the global posterior with Adam
    (learning rate $0.001$) for 300 steps. Every 30 steps, the current model is
    evaluated on the validation set, and the checkpoint with the highest validation
    accuracy is kept. The five resulting models form the final ensemble, whose
    predicted class probabilities are averaged.
\end{itemize}
\subsection{Practical Implementation Considerations}
\label{appx:practical}

\paragraph{Communication efficiency.}
We assume that a scalar is represented with 32 bits by default, so a FedAvg
client uploads its local model $\mathbf W_k$ at a cost of $32d$ bits. A FedFiOT
client additionally uploads its diagonal Fisher $\widetilde{\mathbf F}_k$, which
also has exactly $d$ entries, doubling the uncompressed upload to $64d$ bits. To
match the cost of FedAvg, we apply the same uniform quantization as
FedFisher~\cite{jhunjhunwala2024fedfisher}. For a vector $x$ (one layer of
$\mathbf W_k$ or $\widetilde{\mathbf F}_k$) and $l_q=2^{15}-1$ levels, each
entry is quantized as
\begin{equation}
    [Q(x)]_i
    =
    \|x\|_\infty\,
    \frac{\operatorname{round}\!\left(l_q\,x_i/\|x\|_\infty\right)}{l_q}.
\end{equation}
Each quantized entry takes one of $2l_q+1<2^{16}$ values and is therefore sent
with 16 bits (15 for the magnitude and 1 for the sign), plus 32 bits per layer
for $\|x\|_\infty$. Quantizing both the weights and the Fisher layer-wise gives
a total upload of $2\cdot16d+2\cdot32L=32d+64L$ bits, where $L$ is the number of
layers. For our CNN, $L=5$ and $d\approx0.94$M, so the 320-bit overhead relative
to FedAvg is negligible. The server applies the tempering exponent $\rho$ and
damping $\lambda$ after receiving the uploads, so the hyperparameter sweep of
Appendix~\ref{appx:Hyperparameter} requires no additional communication.

Table~\ref{tab:quantization} compares FedFiOT with full-precision and quantized
uploads. Quantization changes the mean accuracy by at most 1.3 percentage points
in all settings except SVHN with $\alpha=0.1$, where it improves accuracy upon the uncompressed version which can be attributed to an additional regularization effect of compression, as also observed for
FedFisher~\cite{jhunjhunwala2024fedfisher}.

\begin{table}[h]
\centering
\caption{One-shot test accuracy (\%) of FedFiOT with full-precision (32-bit)
and quantized (16-bit) uploads, $K=6$ clients. Mean $\pm$ std over three seeds.}
\label{tab:quantization}
\begin{tabular}{lcccc}
\toprule
Dataset & $\alpha$ & 32-bit & 16-bit & Difference \\
\midrule
\multirow{2}{*}{SVHN}
  & 0.2  & 78.2{\scriptsize$\pm$2.0} & 78.9{\scriptsize$\pm$1.9} & $+0.7$ \\
  & 0.1  & 62.1{\scriptsize$\pm$2.1} & 66.8{\scriptsize$\pm$0.8} & $+4.7$ \\
\midrule
\multirow{3}{*}{CIFAR-10}
  & 0.2  & 51.1{\scriptsize$\pm$0.9} & 52.0{\scriptsize$\pm$0.9} & $+0.9$ \\
  & 0.1  & 44.6{\scriptsize$\pm$1.3} & 43.3{\scriptsize$\pm$1.8} & $-1.3$ \\
\midrule
\multirow{2}{*}{CINIC-10}
  & 0.2  & 45.5{\scriptsize$\pm$2.5} & 45.6{\scriptsize$\pm$1.4} & $+0.1$ \\
  & 0.1  & 38.6{\scriptsize$\pm$0.4} & 39.6{\scriptsize$\pm$0.3} & $+1.0$ \\
\bottomrule
\end{tabular}
\end{table}

\paragraph{Computation efficiency.}
To compute its diagonal Fisher, each client performs one additional pass over
its local data, computing per-example gradients of the observed-label
log-likelihood. This is a modest cost compared with local training, which takes
multiple local epochs in our experiments. On the server, each inner
iteration for layer $\ell$ requires, for every client, building the cost matrix
in $\mathcal O(m_\ell^2 m_{\ell-1})$ time and solving the assignment in
$\mathcal O(m_\ell^3)$ time with the Hungarian algorithm. With at most $S=8$
inner iterations, the merge is cheap
compared with client training.

\paragraph{Computing Infrastructure.} Experiments were conducted on a Linux workstation running Ubuntu 20.04 LTS, equipped with an Intel Xeon E5-2697 v4 CPU @ 2.30\,GHz (36 cores, 72 threads), 384\,GB RAM, and two NVIDIA RTX A6000 GPUs (48\,GB VRAM). Our implementation is based on
PyTorch~2.8 with CUDA~12.8.

\section{Additional Experimental Results}
\subsection{Extreme non-i.i.d.\ setting}
\label{appx:noniid}

We further evaluate one-shot performance under more extreme heterogeneity by
partitioning CIFAR-10 among $K=6$ clients with Dirichlet parameter
$\alpha=0.05$, so that each client's data is concentrated on very few classes.
All other settings follow Section~\ref{sec:experimental_setup}.
Table~\ref{tab:noniid} reports the results.

\begin{table}[h]
\centering
\caption{One-shot test accuracy (\%) on CIFAR-10 with $K=6$ clients and
$\alpha=0.05$. Mean $\pm$ std over three seeds.}
\label{tab:noniid}
\begin{tabular}{lc}
\toprule
Method & Accuracy \\
\midrule
FedAvg            & 25.5{\scriptsize$\pm$1.7} \\
OTFusion          & 27.8{\scriptsize$\pm$2.1} \\
PFNM              & 20.5{\scriptsize$\pm$3.0} \\
RegMean           & 26.6{\scriptsize$\pm$2.6} \\
DENSE             & 26.4{\scriptsize$\pm$0.7} \\
FisherMerge       & 28.0{\scriptsize$\pm$3.0} \\
FedFisher (K-FAC) & 31.9{\scriptsize$\pm$1.1} \\
\midrule
FedFiOT (Diag)    & \textbf{41.4}{\scriptsize$\pm$1.7} \\
\bottomrule
\end{tabular}
\end{table}

FedFiOT outperforms the strongest baseline, FedFisher (K-FAC), by 9.5
percentage points and FedAvg by 15.9 points. The advantage of FedFiOT grows
with heterogeneity: on CIFAR-10, its gap over FedFisher (K-FAC) increases from
3.3 points at $\alpha=0.2$ and 4.7 points at $\alpha=0.1$ to 9.5 points at
$\alpha=0.05$. Moreover, FedFiOT loses only 1.9 points when $\alpha$ decreases
from $0.1$ to $0.05$, whereas FedFisher (K-FAC) loses 6.7 points and PFNM
loses 12.3 points. These results are consistent with our motivation: under
stronger heterogeneity, client models drift further apart, neuron misalignment
becomes more severe, and jointly aligning neurons and weighting parameters by
their Fisher importance becomes more beneficial.

\subsection{Comparisons with Ensemble-based Methods}
\label{appx:ensemble}
\begin{figure*}[h!]
\centering
\includegraphics[width=0.8\textwidth]{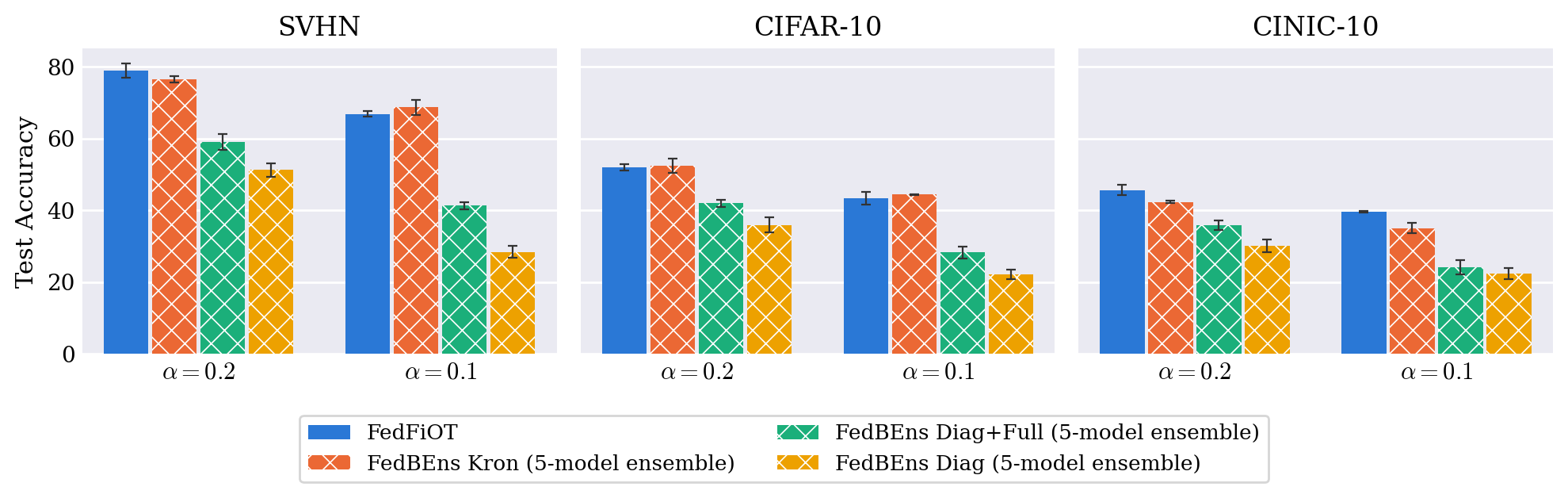}
\caption{One-shot test accuracy (\%) of FedFiOT and FedBEns with six
clients, mean $\pm$ sample standard deviation over three training seeds.
FedFiOT returns a single model; each FedBEns variant returns a five-model
ensemble. FedBEns Kron selects server checkpoints on validation data.}
\label{fig:fedbens}
\end{figure*}

We compare FedFiOT with FedBEns~\cite{talpini2025fedbens}, which returns an
ensemble rather than a single model. Each FedBEns client trains five models,
one from each of five shared random initializations, and fits a Laplace
approximation around each. The server then constructs five global models from
the resulting mixture posteriors (Appendix~\ref{appx:Hyperparameter}). We
evaluate three variants that differ in their curvature approximation: Kron
(Kronecker-factored), Diag+Full (diagonal, with a full covariance for the last
layer), and Diag (diagonal). For Kron, we report the validation-selected
version, which is the strongest. Figure~\ref{fig:fedbens} reports the results.

FedFiOT outperforms FedBEns Diag and Diag+Full in every setting. Against the strongest variant, FedBEns Kron, FedFiOT
is more accurate in three of the six settings (by 2.5 to 4.6 points on SVHN
with $\alpha=0.2$ and on CINIC-10) and less accurate in the other three (by 0.4
to 1.9 points). FedFiOT achieves this comparable accuracy at a substantially
lower cost, summarized in Table~\ref{tab:ensemble_cost}. FedBEns requires
each client to train five models instead of one and to upload all five
models together with their curvature estimates, and it requires five forward
passes for every prediction. In contrast, FedFiOT trains a single model per
client, matches the upload cost of FedAvg, and returns a single model.
\begin{table}[h!]
\centering
\caption{Cost comparison between FedFiOT and FedBEns per client and at
inference, relative to FedAvg.}
\label{tab:ensemble_cost}
\begin{tabular}{lcc}
\toprule
 & FedFiOT & FedBEns \\
\midrule
Models trained per client      & 1 & 5 \\
Upload per client              & $1\times$ FedAvg & $>5\times$ FedAvg \\
Models stored at inference     & 1 & 5 \\
Forward passes per prediction  & 1 & 5 \\
\bottomrule
\end{tabular}
\end{table}
\subsection{Different client initialization}
\label{appx:diff_init}

In the main experiments, all clients start local training from the same broadcast model, which partially
aligns their neurons before merging. Here, each client instead initializes its model independently at
random, as in model merging of independently trained networks~\citep{singh2020model,ainsworth2022git}.
This setting arises when clients hold pre-existing models or when the server cannot coordinate the
initialization. All other settings follow Section~\ref{sec:experimental_setup}. Table~\ref{tab:diff_init} reports the results.

\begin{table}[h]
\centering
\caption{One-shot test accuracy (\%) with independently initialized clients, $K=6$. Mean $\pm$ std over
three seeds.}
\label{tab:diff_init}
\setlength{\tabcolsep}{4pt}
\begin{tabular}{lcccccc}
\toprule
 & \multicolumn{2}{c}{SVHN} & \multicolumn{2}{c}{CIFAR-10} & \multicolumn{2}{c}{CINIC-10} \\
\cmidrule(lr){2-3}\cmidrule(lr){4-5}\cmidrule(lr){6-7}
Method & $\alpha=0.2$ & $\alpha=0.1$ & $\alpha=0.2$ & $\alpha=0.1$ & $\alpha=0.2$ & $\alpha=0.1$ \\
\midrule
FedAvg            & 41.2{\scriptsize$\pm$4.3} & 17.6{\scriptsize$\pm$1.3} & 16.3{\scriptsize$\pm$2.5} & 18.9{\scriptsize$\pm$2.8} & 14.9{\scriptsize$\pm$0.8} & 19.6{\scriptsize$\pm$2.6} \\
OTFusion          & 55.6{\scriptsize$\pm$0.5} & 30.2{\scriptsize$\pm$1.4} & 36.5{\scriptsize$\pm$1.0} & 30.6{\scriptsize$\pm$0.4} & 33.8{\scriptsize$\pm$0.5} & 28.4{\scriptsize$\pm$1.3} \\
WB Fusion         & 59.7{\scriptsize$\pm$2.6} & 34.3{\scriptsize$\pm$0.3} & 38.6{\scriptsize$\pm$1.6} & 27.4{\scriptsize$\pm$1.7} & 35.7{\scriptsize$\pm$0.9} & 25.8{\scriptsize$\pm$1.1} \\
PFNM              & 68.9{\scriptsize$\pm$3.1} & 39.3{\scriptsize$\pm$0.3} & 42.8{\scriptsize$\pm$2.7} & 33.8{\scriptsize$\pm$1.4} & 38.4{\scriptsize$\pm$1.5} & 25.4{\scriptsize$\pm$0.9} \\
RegMean           & 32.0{\scriptsize$\pm$1.6} & 25.4{\scriptsize$\pm$2.6} & 20.8{\scriptsize$\pm$1.4} & 18.2{\scriptsize$\pm$1.1} & 16.7{\scriptsize$\pm$1.1} & 17.8{\scriptsize$\pm$1.1} \\
DENSE             & 65.5{\scriptsize$\pm$0.9} & 47.0{\scriptsize$\pm$1.1} & 34.0{\scriptsize$\pm$2.3} & 28.4{\scriptsize$\pm$0.6} & 26.0{\scriptsize$\pm$1.8} & 22.9{\scriptsize$\pm$1.1} \\
FisherMerge       & 30.8{\scriptsize$\pm$3.0} & 18.9{\scriptsize$\pm$4.6} & 14.3{\scriptsize$\pm$1.7} & 18.1{\scriptsize$\pm$3.6} & 15.1{\scriptsize$\pm$1.7} & 13.2{\scriptsize$\pm$1.2} \\
FedFisher (K-FAC) & 57.3{\scriptsize$\pm$1.4} & 42.6{\scriptsize$\pm$5.5} & 34.2{\scriptsize$\pm$3.4} & 30.6{\scriptsize$\pm$3.2} & 29.3{\scriptsize$\pm$0.5} & 26.8{\scriptsize$\pm$1.1} \\
\midrule
FedFiOT           & \textbf{79.7}{\scriptsize$\pm$0.3} & \textbf{58.4}{\scriptsize$\pm$4.1} & \textbf{52.1}{\scriptsize$\pm$2.0} & \textbf{44.8}{\scriptsize$\pm$1.5} & \textbf{45.2}{\scriptsize$\pm$2.2} & \textbf{38.6}{\scriptsize$\pm$1.4} \\
\bottomrule
\end{tabular}
\end{table}

Without a shared initialization, methods that average parameters coordinate-wise degrade sharply. On
CIFAR-10 and CINIC-10, FedAvg, FisherMerge, and RegMean fall to 13--21\% accuracy, close
to the 10\% of random guessing. FedFisher (K-FAC), the strongest baseline with a shared initialization,
loses 5.2--19.0 percentage points. Neuron-matching methods
are more robust, and PFNM becomes the strongest baseline in four of the six
settings. FedFiOT, in contrast, retains nearly the same accuracy as with a
shared initialization (Table~\ref{tab:oneshot}) in all settings except SVHN
with $\alpha=0.1$. As a result, its margin over the
strongest baseline grows from 3.3--7.6 points with a shared initialization to 6.8--11.4 points. Alignment
is therefore essential once clients do not share an initialization, but alignment alone is not enough:
OTFusion and WB Fusion also align neurons, yet with a Euclidean metric they trail FedFiOT by a significant margin.

\subsection{Fraction of Neurons Permuted per Round}
\label{appx:frac_perm}

\begin{figure}[h]
\centering
\includegraphics[width=0.8\textwidth]{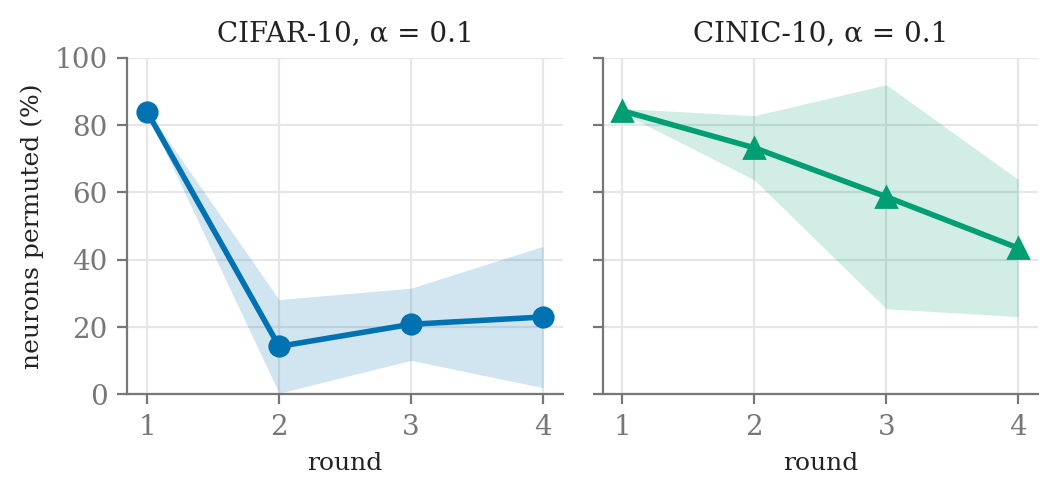}
\caption{Fraction of hidden neurons (\%) permuted by FedFiOT in each round,
$K=6$ clients and $\alpha=0.1$. Lines show the mean over three seeds and
shaded regions show $\pm$ one standard deviation.}
\label{fig:frac_perm}
\end{figure}

To understand how the role of alignment changes across rounds, we record the
fraction of hidden neurons that FedFiOT permutes in each round of the
few-round experiments (Section~\ref{sec:fewround_extension}). Figure~\ref{fig:frac_perm} reports this
fraction.

In the first round, FedFiOT permutes about 84\% of hidden neurons on both
datasets. This indicates that client
models trained from a shared random initialization are substantially
misaligned after local training.

In later rounds, the permuted fraction decreases. On CIFAR-10, it drops
sharply to 14--23\% from the second round on. On CINIC-10, it decreases more
gradually, from 84\% to 44\% by the fourth round.

\section{FedFiOT's Pseudocode}
\label{appx:Psuedocode}
Algorithm~\ref{alg:fedfiot} summarizes the full FedFiOT procedure for one-shot
and few-round FL.

\begin{algorithm}[H]
\caption{FedFiOT for One-shot and Few-round FL}
\label{alg:fedfiot}
\DontPrintSemicolon
\KwIn{
Client datasets $\{\mathcal D_k\}_{k=1}^{K}$;
initial model $\mathbf W^{(0)}$;
communication rounds $T$;
local steps $E$;
maximum inner iterations $S\geq1$;
damping $\lambda>0$;
tempering exponent $\rho\in(0,1]$
}
\KwOut{Global model $\mathbf W^{(T)}$}

\For{$t=0,\dots,T-1$}{
    Select a nonempty client set $\mathcal K^{(t)}$\;
    Broadcast $\mathbf W^{(t)}$ to clients in $\mathcal K^{(t)}$\;

    \ForEach{$k\in\mathcal K^{(t)}$ \textnormal{in parallel}}{
        $\mathbf W_k
        \leftarrow
        \operatorname{LocalTrain}
        (\mathbf W^{(t)},\mathcal D_k,E)$\;
        $\widetilde{\mathbf F}_k
        \leftarrow
        \operatorname{DiagonalEmpiricalFisher}
        (\mathbf W_k,\mathcal D_k)$\;
        Upload quantized versions of $(\mathbf W_k,\widetilde{\mathbf F}_k)$
        to the server\;
    }

    Set $\boldsymbol\pi_k^0\leftarrow I_{m_0}$
    for all $k\in\mathcal K^{(t)}$\ . Set $\widetilde{\mathbf F}_k\leftarrow(\widetilde{\mathbf F}_k)^\rho$ for all $k\in\mathcal K^{(t)}$\;

    \For{$\ell=1,\dots,L$}{
        \tcp{Reorder incoming coordinates using the previous layer}
        \ForEach{$k\in\mathcal K^{(t)}$}{
            \For{$j=1,\dots,m_\ell$}{
                $w_{k,j}^\ell
                \leftarrow
                \boldsymbol\pi_k^{\ell-1}w_{k,j}^\ell$\;
                $\widetilde f_{k,j}^\ell
                \leftarrow
                \boldsymbol\pi_k^{\ell-1}
                \widetilde f_{k,j}^\ell
                (\boldsymbol\pi_k^{\ell-1})^\top$\;
            }
        }

        $w^{\ell,[0]} \leftarrow w_{\mathrm{ref}}^\ell$\tcp*{Largest-client reference or broadcast reference}

        \For{$s=1,\dots,S$}{
            \tcp{Fisher-weighted neuron alignment}
            \ForEach{$k\in\mathcal K^{(t)}$}{
                \eIf{$\ell<L$}{
                    \For{$i,j=1,\dots,m_\ell$}{
                        $\displaystyle
                        \bigl[C_k^{\ell,[s]}\bigr]_{ij}
                        \leftarrow
                        (w_i^{\ell,[s-1]}
                         -w_{k,j}^\ell)^\top
                        (\widetilde f_{k,j}^\ell
                         +\lambda I)
                        (w_i^{\ell,[s-1]}
                         -w_{k,j}^\ell)$\;
                    }
                    $\boldsymbol\pi_k^{\ell,[s]}
                    \leftarrow
                    \operatorname{Hungarian}
                    (C_k^{\ell,[s]})$\;
                }{
                    $\boldsymbol\pi_k^{\ell,[s]}\leftarrow I_{m_\ell}$\;
                }

                \For{$i=1,\dots,m_\ell$}{
                    $\displaystyle
                    \widehat w_{k,i}^\ell
                    \leftarrow
                    \sum_{j=1}^{m_\ell}
                    \pi_{k,ij}^{\ell,[s]}
                    w_{k,j}^\ell$\;
                    $\displaystyle
                    \widehat f_{k,i}^{\ell,\lambda}
                    \leftarrow
                    \sum_{j=1}^{m_\ell}
                    \pi_{k,ij}^{\ell,[s]}
                    \widetilde f_{k,j}^\ell
                    +\lambda I$\;
                }
            }

            \tcp{Fisher-weighted neuron aggregation}
            \For{$i=1,\dots,m_\ell$}{
                $\displaystyle
                w_i^{\ell,[s]}
                \leftarrow
                \left(
                    \sum_{k\in\mathcal K^{(t)}}
                    \widehat f_{k,i}^{\ell,\lambda}
                \right)^{-1}
                \sum_{k\in\mathcal K^{(t)}}
                    \widehat f_{k,i}^{\ell,\lambda}
                    \widehat w_{k,i}^\ell$\;
            }

            \If{$\ell=L$}{
                \textbf{break}\;
            }
            \If{$s>1$ \textnormal{and}
                $\boldsymbol\pi_k^{\ell,[s]}=\boldsymbol\pi_k^{\ell,[s-1]}$
                \textnormal{for all} $k\in\mathcal K^{(t)}$}{
                \textbf{break}\;
            }
        }

        $\mathbf w^{\ell,(t+1)}
        \leftarrow\mathbf w^{\ell,[s]}$\;
        Set $\boldsymbol\pi_k^\ell\leftarrow\boldsymbol\pi_k^{\ell,[s]}$
        for all $k\in\mathcal K^{(t)}$\;
    }
    \tcp{Keep the alignment only if it does not increase the merge objective}
    \If{$\mathcal J^{(t)}\bigl(\mathbf W^{(t+1)},(\boldsymbol\pi_k^\ell)_{k,\ell}\bigr)>\min_{\mathbf W}\mathcal J^{(t)}(\mathbf W,\mathbf I)$}{
        $\mathbf W^{(t+1)}\leftarrow\arg\min_{\mathbf W}\mathcal J^{(t)}(\mathbf W,\mathbf I)$\;
    }
}
\Return{$\mathbf W^{(T)}$}\;
\end{algorithm}

\section{Ablation Study}
\label{appx:ablation}

To isolate the contribution of each component of FedFiOT, we evaluate variants
that remove or replace one component at a time: neuron alignment, the
Fisher-weighted matching cost and Fisher-weighted aggregation. The variants are:
\begin{itemize}
    \item[(A)] \textbf{FedFiOT}: Fisher-weighted matching and aggregation, $S=8$.
    \item[(B)] \textbf{No alignment}: identity permutations, with the same Fisher-weighted aggregation.
    \item[(C)] \textbf{Euclidean matching}: unit weights in the matching cost, with the same Fisher-weighted aggregation.
    \item[(D)] \textbf{Uniform aggregation}: Fisher-weighted matching, with the aligned clients averaged uniformly.
    \item[(E)] \textbf{WB Fusion}: Euclidean matching \emph{and} uniform aggregation, i.e.\ (C) and (D) combined~\citep{akash2022wasserstein}.
\end{itemize}

\begin{table*}[h]
\centering
\caption{Ablation of FedFiOT's components on $K=6$. Test accuracy (\%),
mean $\pm$ std over three seeds.}
\label{tab:ablation}
\small
\setlength{\tabcolsep}{3.5pt}
\begin{tabular}{lccccccc}
\toprule
 & \multicolumn{2}{c}{SVHN} & \multicolumn{3}{c}{CIFAR-10} & \multicolumn{2}{c}{CINIC-10} \\
\cmidrule(lr){2-3}\cmidrule(lr){4-6}\cmidrule(lr){7-8}
Variant & $\alpha=0.2$ & $\alpha=0.1$ & $\alpha=0.2$ & $\alpha=0.1$ & $\alpha=0.05$ & $\alpha=0.2$ & $\alpha=0.1$ \\
\midrule
(A) FedFiOT                & \textbf{78.9}{\scriptsize$\pm$1.9} & \textbf{66.8}{\scriptsize$\pm$0.8} & \textbf{52.0}{\scriptsize$\pm$0.9} & \textbf{43.3}{\scriptsize$\pm$1.8} & \textbf{41.4}{\scriptsize$\pm$1.7} & \textbf{45.6}{\scriptsize$\pm$1.4} & \textbf{39.6}{\scriptsize$\pm$0.3} \\
\midrule
(B) No alignment           & 60.4{\scriptsize$\pm$2.7} & 42.4{\scriptsize$\pm$5.6} & 40.1{\scriptsize$\pm$4.2} & 33.2{\scriptsize$\pm$1.5} & 27.6{\scriptsize$\pm$3.9} & 33.1{\scriptsize$\pm$2.0} & 28.2{\scriptsize$\pm$1.6} \\
(C) Euclidean matching     & 77.3{\scriptsize$\pm$0.2} & 49.9{\scriptsize$\pm$4.7} & 50.0{\scriptsize$\pm$1.3} & 34.4{\scriptsize$\pm$1.8} & 27.7{\scriptsize$\pm$4.1} & 42.2{\scriptsize$\pm$1.2} & 32.0{\scriptsize$\pm$0.5} \\
(D) Uniform aggregation    & 61.3{\scriptsize$\pm$1.7} & 38.1{\scriptsize$\pm$1.0} & 38.2{\scriptsize$\pm$1.9} & 31.7{\scriptsize$\pm$1.2} & 21.1{\scriptsize$\pm$0.7} & 34.4{\scriptsize$\pm$1.0} & 26.2{\scriptsize$\pm$0.9} \\
(E) WB Fusion              & 60.8{\scriptsize$\pm$1.0} & 34.7{\scriptsize$\pm$2.3} & 40.3{\scriptsize$\pm$1.2} & 29.0{\scriptsize$\pm$1.1} & 18.3{\scriptsize$\pm$0.7} & 36.4{\scriptsize$\pm$1.1} & 25.4{\scriptsize$\pm$0.8} \\
\bottomrule
\end{tabular}
\end{table*}

\textbf{Each component of the merge is necessary.} Removing alignment (B) reduces accuracy by 10--24
points, and replacing Fisher-weighted with uniform aggregation (D) by 11--29 points. Replacing the
Fisher-weighted matching cost with a Euclidean one (C) costs 2--3 points at $\alpha=0.2$ but 8--17 points
at $\alpha\le0.1$: the more heterogeneous the clients, the more it matters that neurons are matched under
the same metric in which they are averaged. Removing both Fisher weightings yields WB
Fusion (E), the weakest variant in four of the seven settings. Hence neither alignment nor Fisher-weighted
aggregation alone explains FedFiOT's gains, but using the Fisher metric consistently in both steps does.

\section{Numerical Example}
\label{appx:example}

We illustrate the alternating updates of Section~\ref{sec:gen_proc} on a
hidden layer with three neurons, merged from two clients. Each neuron has
one scalar incoming weight, and we suppress $\ell$. Let
\[
\mathbf w^{[0]}=\mathbf w_1=(0,4,8)^\top,
\qquad
\mathbf w_2=(10,2,6)^\top.
\]
Client~2 learns the same three features as client~1, shifted by $+2$ and
stored in a different order: its first neuron corresponds to global
neuron~3, its second to global neuron~1, and its third to global neuron~2.
Given $\lambda=1$ and Fisher values $(2,0,2)$ for client~1 and $(0,0,2)$ for
client~2, the damped Fisher values are $(3,1,3)$ and $(1,1,3)$, respectively.

\paragraph{First alignment step.}
Equation~\eqref{eq:fisher_ground_cost} gives
\[
C_1^{[1]}=
\begin{pmatrix}0&16&192\\48&0&48\\192&16&0\end{pmatrix},
\qquad
C_2^{[1]}=
\begin{pmatrix}100&4&108\\36&4&12\\4&36&12\end{pmatrix},
\]
where rows index global neurons and columns index client neurons.
Client~1 coincides with $\mathbf w^{[0]}$, so it retains its order at zero
cost. For client~2, write an alignment as the client neurons assigned to
global neurons $1,2,3$. The identity $(1,2,3)$ costs $116$; the swaps
$(2,1,3)$, $(3,2,1)$, and $(1,3,2)$ cost $52$, $116$, and $148$; and the
cycles $(2,3,1)$ and $(3,1,2)$ cost $20$ and $180$.
The assignment~\eqref{eq:alignment_update} therefore selects the cycle
$(2,3,1)$,
\[
\boldsymbol\pi_2^{[1]}=
\begin{pmatrix}0&1&0\\0&0&1\\1&0&0\end{pmatrix},
\]
which recovers the true correspondence; every swap leaves two neurons
mismatched. Client~2's Fisher values move with its neurons,
so by~\eqref{eq:aligned_neuron_quantities} its aligned weights and damped
Fisher values are $(2,6,10)$ and $(1,3,1)$, respectively.

\paragraph{First aggregation step.}
Equation~\eqref{eq:closed_form} gives
\[
\mathbf w^{[1]}=
\begin{pmatrix}
(3\cdot0+1\cdot2)/(3+1)\\
(1\cdot4+3\cdot6)/(1+3)\\
(3\cdot8+1\cdot10)/(3+1)
\end{pmatrix}
=
\begin{pmatrix}0.5\\5.5\\8.5\end{pmatrix}.
\]
Each merged weight moves toward the client with greater precision for that
neuron: toward client~1 for global neurons~1 and~3, and toward client~2 for
global neuron~2.

\paragraph{Second alignment step.}
With $\mathbf w^{[1]}$ as the reference, the cheapest assignments are again
the identity for client~1 (cost $3.75$; next best $39.75$) and the cycle
$(2,3,1)$ for client~2 (cost $5.25$; next best $41.25$). Since no alignment
changes, the layer merge stops after two iterations. The objective decreases
from $\mathcal J^{[0]}=116$, the cost of the identity alignments at
$\mathbf w^{[0]}$, to $\mathcal J^{[1]}=3.75+5.25=9$, as
Theorem~\ref{thm:fedfiot_stability} guarantees.

\paragraph{Comparison.}
Table~\ref{tab:numerical_example} compares FedFiOT with merges that omit
alignment, Fisher weighting, or both. Without alignment, global neuron~1
combines client~1's weight $0$ with client~2's weight $10$, although the two
neurons encode different features. Fisher merging without alignment
attains $\mathcal J=83$. In this single-layer example the full-model
objective equals $\mathcal J$, so the identity-aligned comparison at the end
of the merge keeps the aligned result.

\begin{table}[h]
\centering
\caption{Merged layer and objective $\mathcal J$ for the three-neuron
example. Methods without alignment use identity alignments for both
clients.}
\label{tab:numerical_example}
\begin{tabular}{lccc}
\toprule
Merge & Alignment & $\mathbf w^\top$ & $\mathcal J$ \\
\midrule
Average by index & identity & $(5,\,3,\,7)$ & $108$ \\
Fisher merging & identity & $(2.5,\,3,\,7)$ & $83$ \\
Average of aligned weights & $(2,3,1)$ & $(1,\,5,\,9)$ & $12$ \\
FedFiOT & $(2,3,1)$ & $(0.5,\,5.5,\,8.5)$ & $9$ \\
\bottomrule
\end{tabular}
\end{table}






\clearpage
\appendix
\thispagestyle{empty}

\end{document}